\documentclass[11pt]{article}
\ifdefined\pdfminorversion\pdfminorversion=7\fi

\PassOptionsToPackage{table}{xcolor}
\usepackage[preprint]{acl}
\usepackage{times}
\usepackage{latexsym}
\usepackage[T1]{fontenc}
\usepackage[utf8]{inputenc}
\usepackage{microtype}
\usepackage{inconsolata}

\usepackage{amsmath,amssymb,mathtools}
\usepackage{graphicx}
\usepackage{booktabs,tabularx,array}

\usepackage{tikz}
\usetikzlibrary{arrows.meta,positioning,calc}

\usepackage{enumitem}
\usepackage{algorithm,algpseudocode}
\usepackage{float,needspace}

\usepackage{placeins}


\setcitestyle{numbers,square}

\newcommand{\evalresult}[2]{\mbox{#1{\scriptsize\,$\pm$\,#2}}}

\newcommand{\sg}{\operatorname{stopgrad}}
\newcommand{\clip}{\operatorname{clip}}

\newcommand{\Cov}{\operatorname{Cov}}

\definecolor{ResultBest}{HTML}{FFF0C9}
\definecolor{ResultSecond}{HTML}{E3EDFC}
\definecolor{ActorBlue}{HTML}{245C99}
\definecolor{AnalyzerGreen}{HTML}{176D65}
\definecolor{CreditOrange}{HTML}{B56B16}
\definecolor{UpdatePurple}{HTML}{70579C}
\definecolor{Ink}{HTML}{26333E}

\usepackage[most]{tcolorbox}

\newtcolorbox{promptpanel}[1]{%
  enhanced,
  unbreakable,
  colback=white,
  colframe=ActorBlue!30!white,
  colbacktitle=ActorBlue!6!white,
  coltitle=ActorBlue!85!black,
  title={#1},
  fonttitle=\normalfont\small\bfseries,
  fontupper=\normalfont\small,
  boxrule=0.5pt,
  arc=1.5mm,
  left=8pt,
  right=8pt,
  top=6pt,
  bottom=6pt,
  toptitle=5pt,
  bottomtitle=5pt,
  before skip=7pt,
  after skip=7pt,
  before upper={%
    \raggedright
    \setlength{\parindent}{0pt}%
    \setlength{\parskip}{2pt}%
  }%
}

\newcommand{\promptslot}[1]{%
  {\ttfamily\color{ActorBlue!85!black}#1}%
}

\title{ECOS: Environment-Calibrated On-Policy Self-Distillation
for Language Agents}

\author{%
  Chenhao Zhang$^{1}$ \quad Zhongan Wang$^{2}$ \quad
  Mingxuan Li$^{1}$ \quad
  \textbf{Hehe Fan}$^{2}$ \\
  \textbf{Zheng Zhu}$^{3}$ \quad 
  \textbf{Ming Li}$^{1}$\thanks{Corresponding author: Ming Li (ming.li@u.nus.edu).} \\
  {\normalfont\normalsize $^{1}$School of Artificial Intelligence, The Chinese University of Hong Kong, Shenzhen} \\
  {\normalfont\normalsize $^{2}$College of Artificial Intelligence, Zhejiang University} \\
  {\normalfont\normalsize $^{3}$GigaAI}%
}
\date{}
\begin{document}
\maketitle

% ---------------- Abstract ----------------
\begin{abstract}
On-policy self-distillation lets a language agent act as its own Teacher by conditioning on privileged context such as skills written from its own trajectories. Its guidance, however, is weighted by how much this context shifts token likelihoods rather than by whether it improves behavior, so errors in self-written skills return through later updates. We propose environment-calibrated on-policy self-distillation, which uses environmental outcomes to decide how much to trust the Teacher's current guidance and which self-written skills to reinforce. Command-level advantages, estimated by comparing commands executed at similar observations, scale each Teacher preference and reward reflections and skills whose induced preferences favor higher-advantage commands. Without skills at inference, 3B and 7B agents reach \(90.5\%\) and \(93.3\%\) macro success on seen ALFWorld tasks and \(75.5\%\) and \(80.5\%\) success on WebShop. On unseen ALFWorld tasks, they outperform GRPO under the same protocol by \(11.0\) and \(7.6\) points. Supervising only executable commands rather than the Actor's full response raises 3B WebShop success from \(52.3\%\) to \(75.0\%\).
\end{abstract}

% ---------------- Main text ----------------
\section{Introduction}
\label{sec:intro}

Large language model agents solve multi-step tasks by issuing commands and reading the resulting observations~\cite{stateopd,self}. Reinforcement learning (RL) trains such agents from environmental rewards, but sparse and delayed rewards say little about which intermediate decisions produced an outcome~\cite{rudder,gigpo}. On-policy distillation (OPD) adds token-level Teacher preferences along trajectories generated by the policy itself~\cite{gkd}. On-policy self-distillation (OPSD) obtains these preferences without a separate Teacher model: the policy acts as its own Teacher when its context contains privileged information, such as feedback or skills summarized from earlier attempts~\cite{opsd,sdpo,skillsd}. When a model generates skills from its own trajectories, every update supplies supervision for subsequent learning. This creates an iterative learning loop. The performance of this loop relies on whether the self-generated supervision is reliable.

In standard OPSD, however, nothing checks whether this supervision is correct: each sampled token is weighted by how much the privileged context changes its likelihood~\cite{opsd}, whether or not the change favors better behavior. This weighting can fail for agents in three ways. The task may exceed what the Teacher knows, and an erroneous Actor prefix may steer a capable Teacher toward the same mistake~\cite{criticopd,manyfaces}. Teacher reasoning may also rely on privileged information that the Actor never observes, so imitating it can weaken reasoning grounded in the Actor's own observations~\cite{privilegedcontext}. When the policy writes its own skills, a mistaken reflection can enter a skill and return through the next policy update.

Calibration methods regulate the local signal, align decision contexts, or tie Teacher weights to returns and policy influence~\cite{sdar,pcsd,alignopsd,adrs,icsd}. Their assessment, however, does not reach the process that generates the guidance. Methods that adapt this process~\cite{grsd,self,seed,sdzero} do not train it on whether the resulting guidance favors commands with better environmental outcomes. This raises a question for skill-based self-distillation: can a single environment-grounded criterion decide how much to trust the current Teacher and also train the analysis that builds the next one?

\begin{figure*}[t]
\centering
\includegraphics[width=\linewidth]{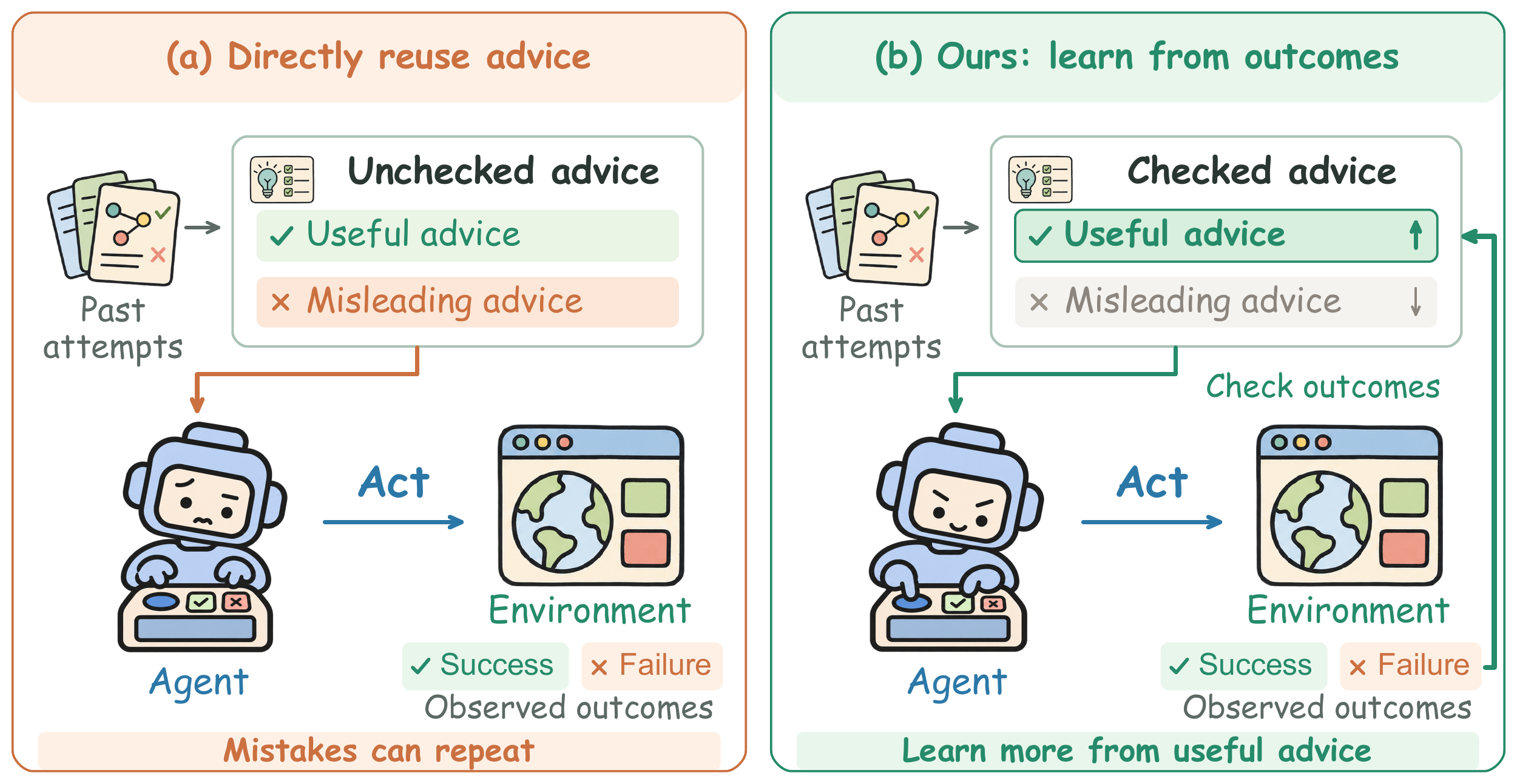}
\caption{\textbf{Environmental outcomes help check an agent's own advice.} \textbf{(a)} Self-generated advice can carry useful lessons and mistakes into later training. \textbf{(b)} Our method uses observed outcomes to give useful advice more influence and conflicting advice less influence. These outcomes also train the generation of future advice. The comparison concerns how advice is assessed during training.}
\label{fig:motivation}
\end{figure*}

Our method uses command-level environmental advantages both to weight current Teacher guidance and to train the analysis that produces the next Teacher. Figure~\ref{fig:motivation} illustrates the central contrast: using environmental outcomes to check self-generated advice. After supervised initialization on externally annotated reflections and skills, a single model acts in the environment (Actor), writes reflections and a consolidated skill for each trajectory group (Analyzer), and serves as the Teacher when conditioned on that skill. The advantages compare commands executed at similar observations and therefore do not depend on the Teacher. The Analyzer earns credit when its skill induces preferences for higher-advantage commands, so the Teacher's task knowledge is learned rather than assumed.

In the Actor update, each Teacher preference is scaled by its agreement with the advantage; guidance misled by an erroneous prefix loses weight without being reversed. Direct Teacher supervision covers only executable commands, so the Actor never imitates reasoning that depends on the skill, and RL trains the full response.

We evaluate on ALFWorld and WebShop with Qwen2.5-3B and 7B backbones. Across three evaluation seeds, the trained agents without skills or online analysis at inference reach $90.5\%$ and $93.3\%$ macro success on seen ALFWorld tasks and $75.5\%$ and $80.5\%$ success on WebShop. On unseen ALFWorld tasks, they outperform GRPO under the same protocol by $11.0$ and $7.6$ points. In 3B ablations, supervision scope has the largest effect: extending Teacher supervision from commands to the full response lowers WebShop success from $75.0\%$ to $52.3\%$, and removing environmental feedback from both updates lowers ALFWorld success by $6.5$ points. At the end of training, only $47$--$53\%$ of the raw Teacher weight agrees with trajectory outcomes, and modulation raises this share to $57$--$74\%$.
Our contributions are as follows:
\begin{enumerate}
    \item We formulate skill-based self-distillation as a closed loop in which the policy writes the skills that condition its own Teacher, and argue that one environmental criterion should both weight current Teacher guidance and train the analysis that produces future guidance.
    \item We instantiate this criterion with command-level advantages: they modulate Teacher preferences in the Actor update and reward reflections and skills in the Analyzer update, while Teacher supervision is restricted to executable commands.
    \item Experiments on ALFWorld and WebShop with 3B and 7B backbones show that the trained agents outperform GRPO on unseen tasks without using skills at inference, and ablations show that command-only supervision and environmental feedback each contribute.
\end{enumerate}

\Needspace{6\baselineskip}
\section{Related Work}
\label{sec:related}

\textbf{Environmental credit for agent learning.}
Outcome-based RL evaluates complete trajectories, whereas credit assignment must distinguish the decisions that produced their outcomes~\cite{ppo,deepseekmath}. Return decomposition assigns delayed outcome credit to earlier decisions~\cite{rudder}. For interactive agents, grouping repeated or behaviorally similar visits makes local alternatives comparable~\cite{gigpo,bipace}; propagating consequences through shared state potentials incorporates their downstream effects~\cite{pgpo}. These comparisons can identify useful commands within failed trajectories. We use environmental command credit to evaluate whether Teacher guidance favors useful decisions.

\textbf{On-policy distillation and experience-based self-improvement.}
Experience-based self-improvement introduces a complementary problem: how should previous interactions become reliable guidance for later behavior? Reflection memories and reusable insights retain experience in context~\cite{reflexion,expel}; iterative revision and trained self-correction modify subsequent responses or policy behavior~\cite{selfrefine,score}. OPD and skill-conditioned OPSD transfer guidance along policy-generated sequences~\cite{gkd,opsd,skillsd}. Shared Actor and Analyzer parameters let the policy and its analysis evolve together, as in SEED~\cite{seed}. Errors in that analysis can nevertheless re-enter later updates, and models may struggle to recognize errors in their own feedback without reliable external evidence~\cite{intrinsic,selfincorrect}. We further optimize the reflections and skills according to the environmental value of the Teacher preferences they produce.

\textbf{Calibration of self-generated supervision.}
Calibration concerns what guidance is transferred and what evidence justifies its influence. Decision-level semantic rewards can weight a full response~\cite{actionopd}; our direct Teacher loss instead covers only executable command content. Local signal regulation, step-aware supervision, and aligned decision contexts control where and how strongly Teacher preferences enter learning~\cite{sdar,pcsd,stepopsd,alignopsd}. Environmental returns and objective influence provide criteria for judging their usefulness~\cite{adrs,icsd}. For training the source of guidance, GRSD evaluates reflection quality through external assessment~\cite{grsd}. Our R/S credit instead evaluates a skill through agreement between its induced Teacher preferences and environmental command advantages.

\Needspace{7\baselineskip}
\section{Method}
\label{sec:method}

Our method jointly trains an acting policy (Actor) and a trajectory-analysis role (Analyzer). Direct Teacher supervision covers executable commands, and environmental command advantages determine its strength. Agreement between these advantages and skill-induced preferences also provides training credit for reflection and skill generation.

\begin{figure*}[t]
\centering
\includegraphics[width=\linewidth]{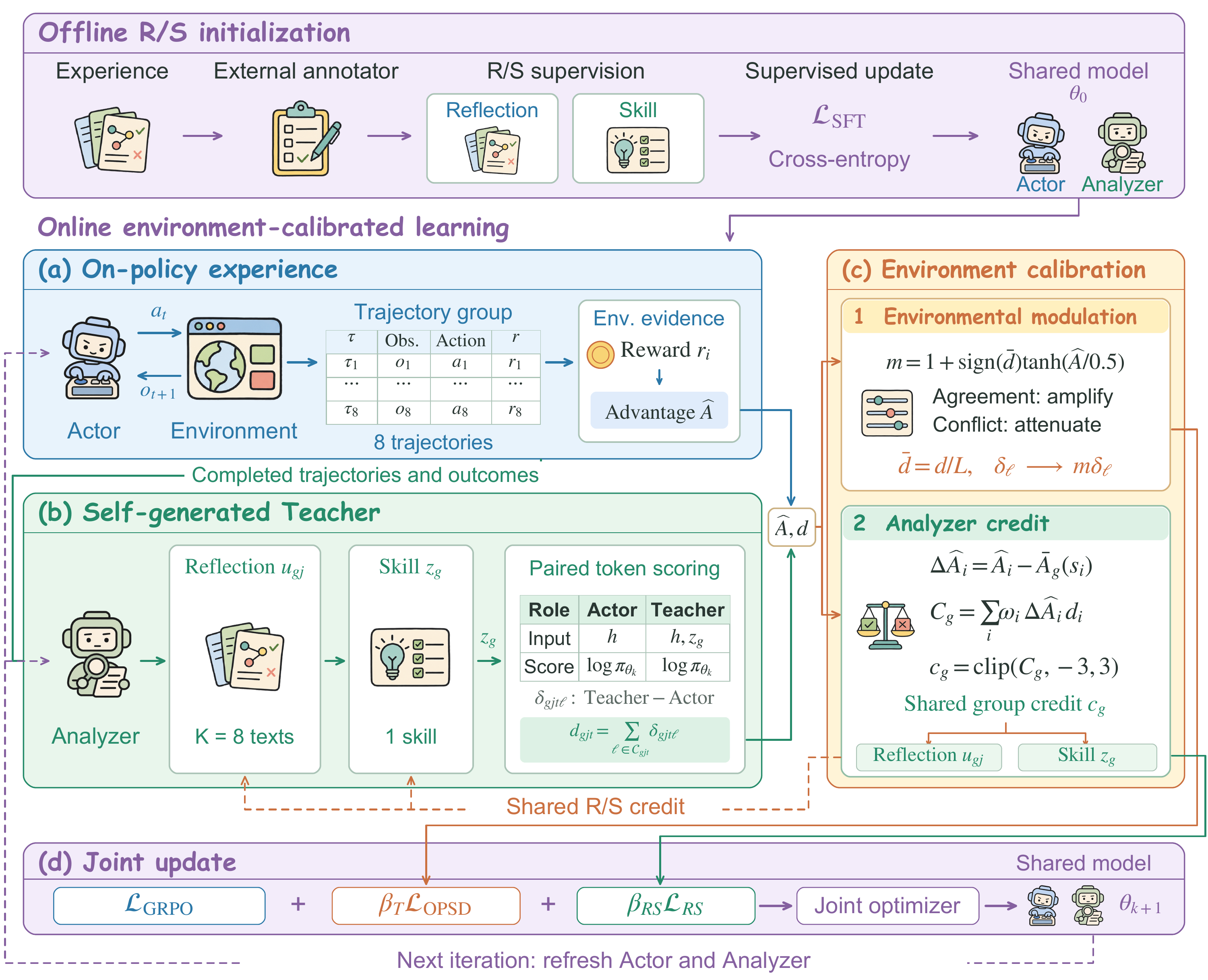}
\caption{\textbf{Environmental feedback connects acting with Teacher construction.} Offline annotations initialize reflection and skill generation through SFT. Online, reflections $u_{gj}$ produce a skill $z_g$, which conditions the pre-update policy to score Actor-sampled tokens. Token gaps $\delta$ are summed into raw command gaps $d$; the mean $\bar d=d/L$ determines a shared modulation multiplier, which scales each command token's own gap. Environmental advantages calibrate this guidance and supply shared credit $c_g$ for reflection and skill generation. Panel (a) omits group and trajectory indices: $o_{t+1}$ is the observation returned after action $a_t$.}
\label{fig:loop}
\end{figure*}

As shown in Figure~\ref{fig:loop}, offline supervised fine-tuning (SFT) initializes reflection and skill generation in the model shared by the Actor and Analyzer. During online training, the Actor collects trajectory groups, and the Analyzer generates one reflection per trajectory and one consolidated skill per group. The skill conditions the pre-update model to form the Teacher. Joint Actor and Analyzer optimization updates the shared parameters for the next iteration.

\subsection{Preliminaries and shared-policy setup}
\label{sec:setup}

Let $x_g$ denote task $g$ and $\theta_k$ the shared model parameters at iteration $k$. Each update collects $G$ task groups, with $K=8$ trajectories $\{\tau_{gj}\}_{j=1}^{K}$ sampled from the Actor $\pi_{\theta_k}$. At interaction step $t$, the Actor generates response $y_{gjt}$ from history $h_{gjt}$. Its executable command specifies action $a_{gjt}$, after which the environment returns observation $o_{gj,t+1}$. We use $\ell$ for a response-token index.

GRPO derives group-relative advantages $\widehat A^{\mathrm{RL}}_{gj}$ from recorded episode rewards~\cite{deepseekmath}. Let $\mathcal B_A$ contain the valid generated Actor-token indices $(g,j,t,\ell)$. The likelihood ratio is
\begin{equation}
    \rho_{gjt\ell}(\theta)
    =
    \frac{
      \pi_{\theta}(y_{gjt\ell}\mid h_{gjt},y_{gjt,<\ell})
    }{
      \pi_{\theta_k}(y_{gjt\ell}\mid h_{gjt},y_{gjt,<\ell})
    }.
    \label{eq:actor-ratio}
\end{equation}
The clipped policy component of the GRPO objective is
\begin{equation}
\begin{aligned}
    &\mathcal L_{\mathrm{GRPO}}^{\mathrm{policy}}(\theta)
    =
    -\frac{1}{|\mathcal B_A|}
    \sum_{(g,j,t,\ell)\in\mathcal B_A}\\
    &\quad\min\!\left\{
    \begin{aligned}
        &\rho_{gjt\ell}(\theta)
          \operatorname{sg}(\widehat A^{\mathrm{RL}}_{gj}),\\
        &\operatorname{clip}
          \bigl(\rho_{gjt\ell}(\theta),1-\epsilon,1+\epsilon\bigr)
          \operatorname{sg}(\widehat A^{\mathrm{RL}}_{gj})
    \end{aligned}
    \right\},
\end{aligned}
\label{eq:grpo-policy}
\end{equation}
where $\epsilon=0.2$ and $\operatorname{sg}$ denotes stop-gradient~\cite{ppo}. This token-mean objective trains the complete generated response, including reasoning. We use $\mathcal L_{\mathrm{GRPO}}$ for this policy loss plus a frozen-reference KL penalty with coefficient $0.01$ and minus an entropy bonus with coefficient $0.001$. Appendix~\ref{app:implementation} gives the optimization settings.

The Analyzer uses the same parameter snapshot under reflection and skill-generation prompts, denoted by $R_{\theta_k}$ and $S_{\theta_k}$. Reflection generation produces one text $u_{gj}$ from each trajectory and its recorded outcome $r_{gj}$. Skill generation consolidates these reflections into one text $z_g$:
\begin{align}
    u_{gj}
    &\sim
    R_{\theta_k}(\cdot\mid x_g,\tau_{gj},r_{gj}),\\
    z_g
    &\sim
    S_{\theta_k}
    \bigl(\cdot\mid x_g,\{u_{gj},r_{gj}\}_{j=1}^{K}\bigr).
\end{align}
For task group $g$, the skill-conditioned Teacher is the pre-update policy evaluated with $z_g$:
\begin{equation}
    \pi_{T,k}(\cdot\mid h_{gjt},y_{gjt,<\ell})
    \equiv
    \pi_{\theta_k}
    (\cdot\mid h_{gjt},z_g,y_{gjt,<\ell}).
    \label{eq:teacher-definition}
\end{equation}
Both branches score the same Actor-sampled token under the same reasoning prefix. Their token-level log-probability gap is
\begin{equation}
\begin{aligned}
    \delta_{gjt\ell}
    ={}&
    \log\pi_{T,k}
    (y_{gjt\ell}\mid h_{gjt},y_{gjt,<\ell})\\
    &-
    \log\pi_{\theta_k}
    (y_{gjt\ell}\mid h_{gjt},y_{gjt,<\ell}).
\end{aligned}
\label{eq:gap}
\end{equation}

In the sampled-token policy-gradient formulation of OPSD, the uncalibrated coefficient is the likelihood gap itself~\cite{gkd,opsd}:
\begin{equation}
    w^{\mathrm{base}}_{gjt\ell}
    =\delta_{gjt\ell}.
    \label{eq:uncalibrated}
\end{equation}
This coefficient measures how privileged context changes token support. It does not explicitly calibrate that support using the environmental value of the corresponding command.

\subsection{Environmental command credit}
\label{sec:environmentcredit}

GRPO assigns the same trajectory-level advantage to every command in an episode, so it cannot distinguish useful decisions within a failed trajectory. We estimate a separate command-level advantage by comparing commands executed at similar observations.

Commands are compared within states formed by representation-based observation grouping~\cite{bipace}. Within each task group, we use frozen-reference embeddings and require identical admissible-action sets. A visit joins the nearest normalized-centroid cluster within cosine distance $0.1$, or creates a new cluster. Recorded commands connect these clusters through their observed successor states and rewards.

\begin{figure*}[tbp]
\centering
\includegraphics[width=\textwidth]{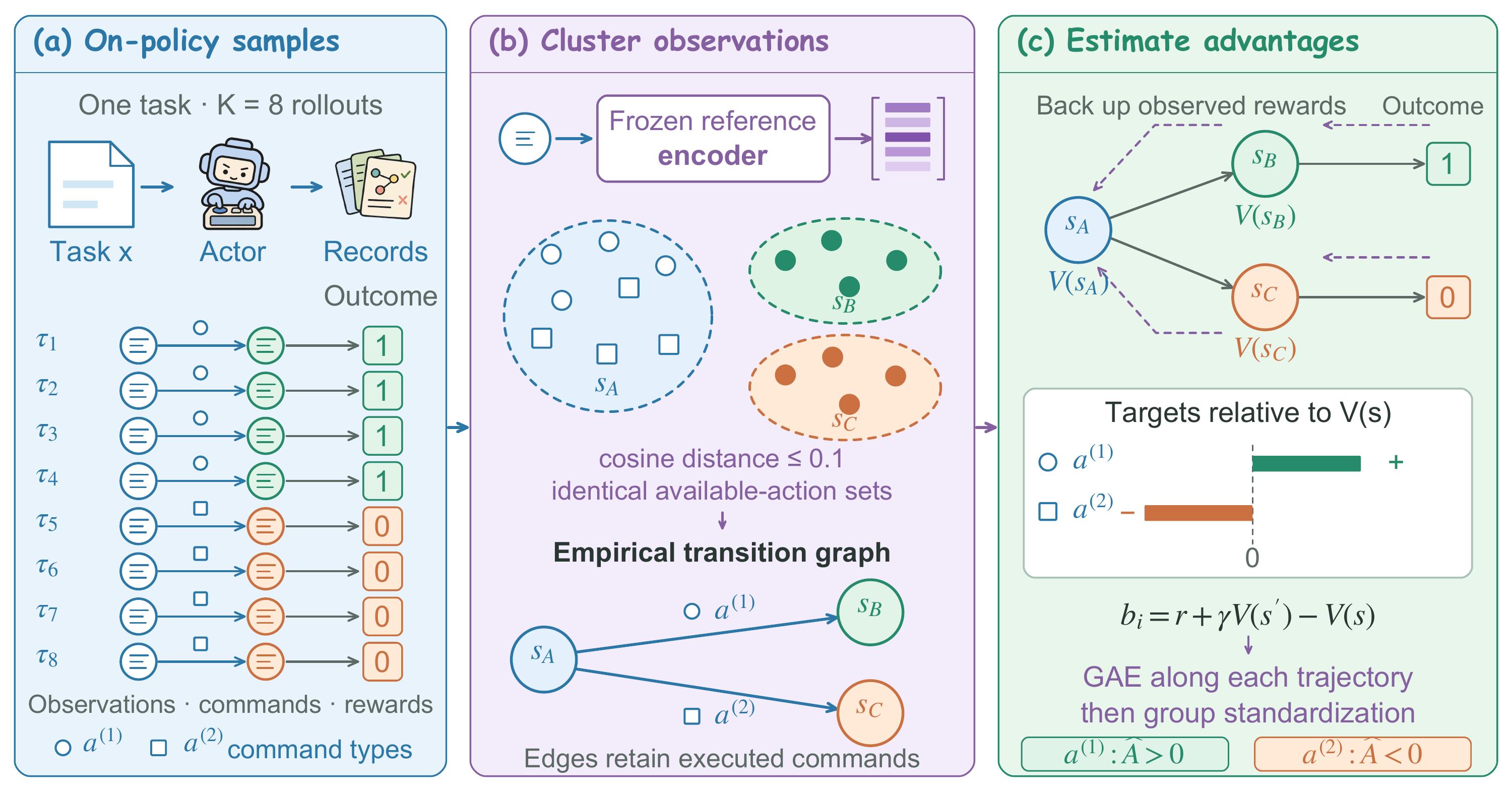}
\caption{\textbf{An empirical transition graph provides environmental command credit.} Recorded observations are grouped into comparable state clusters, and executed commands connect them through observed transitions. Bellman evaluation estimates state values; temporal-difference residuals then enter GAE and group standardization to obtain command advantages. Colors identify observation clusters.}
\label{fig:modulation}
\end{figure*}

For transition $i$, let $s_i$ and $s_i^+$ denote its current and successor clusters, with transition reward $r_i$. We initialize $V^{(0)}(s)$ from the mean observed return-to-go and apply empirical Bellman backups~\cite{gatedbepo}:
\begin{equation}
    V^{(n+1)}(s)
    =
    \frac{1}{|\mathcal V_s|}
    \sum_{i\in\mathcal V_s}
    \bigl[r_i+\gamma V^{(n)}(s_i^+)\bigr],
    \label{eq:valuebackup}
\end{equation}
where $\mathcal V_s$ contains the recorded visits to $s$ and $\gamma=0.95$. Terminal and horizon successors have value zero. The backups propagate observed downstream consequences into the values used for local comparisons.

The temporal-difference residual is
\begin{equation}
    b_i
    =
    r_i+\gamma V(s_i^+)-V(s_i).
\end{equation}
Along each recorded trajectory, generalized advantage estimation~\cite{gae} propagates these residuals:
\begin{equation}
    B_{gjt}
    =
    b_{gjt}
    +\gamma\lambda_{\mathrm{GAE}}B_{gj,t+1},
    \qquad
    \lambda_{\mathrm{GAE}}=0.8,
    \label{eq:envgae}
\end{equation}
with a zero tail value. Group standardization yields the auxiliary command advantage $\widehat A_{gjt}$. It supplies Teacher modulation and Analyzer credit, whereas $\widehat A^{\mathrm{RL}}_{gj}$ supplies the primary GRPO update. Appendix~\ref{app:advantage} gives the solver and normalization details.

\subsection{Environment-calibrated self-distillation}
\label{sec:behavior}

Privileged skills can support executable decisions even when their associated reasoning targets are unavailable to the Actor. We therefore restrict direct Teacher supervision to command content. Let $\mathcal C_{gjt}$ contain the positions in $y_{gjt}$ that encode the executable command for action $a_{gjt}$, excluding delimiter tags, and let $L_{gjt}=|\mathcal C_{gjt}|$. We retain the raw summed gap for R/S credit:
\begin{equation}
    d_{gjt}
    =
    \sum_{\ell\in\mathcal C_{gjt}}\delta_{gjt\ell}.
    \label{eq:commandgap}
\end{equation}
The mean command gap used to compute the modulation multiplier is
\begin{equation}
    \bar d_{gjt}
    =\frac{1}{L_{gjt}}
    \sum_{\ell\in\mathcal C_{gjt}}\delta_{gjt\ell}
    =\frac{d_{gjt}}{L_{gjt}},
    \label{eq:mean-commandgap}
\end{equation}
with $\bar d_{gjt}=0$ for empty command spans.

Environmental credit and the mean command gap determine a shared modulation multiplier:
\begin{equation}
\begin{aligned}
    m_{gjt}
    &=
    1+\alpha_m\operatorname{sign}(\bar d_{gjt})
    \tanh\!\left(\frac{\widehat A_{gjt}}{\kappa_m}\right),\\
    w_{gjt\ell}
    &=
    q_g\,\mathbb{1}[\ell\in\mathcal C_{gjt}]
    \,m_{gjt}\delta_{gjt\ell},
\end{aligned}
\label{eq:modulation}
\end{equation}
with $\alpha_m=1$ and $\kappa_m=0.5$. The multiplier lies in $[0,2]$. Matching signs of $\bar d_{gjt}$ and $\widehat A_{gjt}$ strengthen the guidance, and conflicting signs attenuate it without reversing individual token-gap directions. When $\widehat A_{gjt}=0$, $m_{gjt}=1$ recovers the unmodulated token gaps. Each command shares this multiplier across its content tokens, while each token retains its own gap.

The group mask $q_g$ is one in ALFWorld. In WebShop, it is zero when all eight trajectories succeed and one otherwise. The resulting weights combine command-only supervision, a shared command-level multiplier, and token-specific gaps.

For likelihood ratio $\rho$ and signed weight $w$, define the clipped token loss
\begin{equation}
\begin{aligned}
    &\mathcal J_{\mathrm{clip}}(\rho;w)
    =\\
    &\quad-\min\!\left\{
    \begin{aligned}
       &\rho\,\operatorname{sg}(w),\\
       &\operatorname{clip}(\rho,1-\epsilon,1+\epsilon)
       \operatorname{sg}(w)
    \end{aligned}
    \right\}.
\end{aligned}
    \label{eq:teacherloss}
\end{equation}
Using the Actor token-mean reduction, the calibrated OPSD objective is
\begin{equation}
\begin{aligned}
    &\mathcal L_{\mathrm{OPSD}}(\theta)
    =
    \frac{1}{|\mathcal B_A|}
    \sum_{(g,j,t,\ell)\in\mathcal B_A}\\
    &\qquad\mathcal J_{\mathrm{clip}}
    \bigl(\rho_{gjt\ell}(\theta);w_{gjt\ell}\bigr).
\end{aligned}
    \label{eq:opsd-objective}
\end{equation}
The weights select command content, while the primary GRPO objective continues to train the complete response. Appendix~\ref{app:signal} derives the modulation properties.

\subsection{Credit for reflection and skill generation}
\label{sec:credit}

Command calibration evaluates a given Teacher signal. Training the Analyzer additionally requires evaluating the reflections and skill that produced it. We assign credit according to whether the skill shifts support toward higher-advantage commands at comparable observations. The reflection and skill prompts are provided in Appendix~\ref{app:implementation}.

Within task group $g$, let $\mathcal I_g$ contain the deduplicated transitions with valid command spans, and write $i=(j,t)$. Let $N_g$ be the number of represented trajectories and $L_{gj}$ the number of eligible transitions in trajectory $j$. We use
\begin{equation}
    \omega_i
    =
    \frac{1}{N_gL_{gj}},
    \qquad
    \overline A_g(s)
    =
    \frac{
      \sum_{i\in\mathcal I_g:s_i=s}
      \omega_i\widehat A_i
    }{
      \sum_{i\in\mathcal I_g:s_i=s}\omega_i
    }.
    \label{eq:center}
\end{equation}
Each trajectory receives equal total weight, divided among its eligible transitions. State centering gives
$\Delta\widehat A_i=\widehat A_i-\overline A_g(s_i)$.
This comparison measures relative command value within a cluster rather than differences in the average advantage of visited states.

The raw and bounded group credits are
\begin{equation}
    C_g
    =
    \sum_{i\in\mathcal I_g}
    \omega_i\,\Delta\widehat A_i\,d_i,
    \qquad
    c_g
    =
    \operatorname{clip}(C_g,-3,3).
    \label{eq:credit}
\end{equation}
Here $C_g$ is a weighted within-state covariance between environmental advantage and the raw summed Teacher preference $d_i$. Positive credit reinforces analysis whose Teacher relatively supports higher-advantage commands; negative credit discourages the reverse association.

Because the skill consolidates the full reflection group, all $K$ reflections and the skill receive the same reward $c_g$. For an analysis response $v$ generated under prompt $p_v$, define
\begin{equation}
    \rho^v_\ell(\theta)
    =
    \frac{
      \pi_\theta(v_\ell\mid p_v,v_{<\ell})
    }{
      \pi_{\theta_k}(v_\ell\mid p_v,v_{<\ell})
    }.
\end{equation}
Let $\mathcal M_v$ contain its valid generated-token positions. The response-sequence mean loss uses the same clipped form:
\begin{equation}
    \mathcal P(v;c_g)
    =
    \frac{1}{|\mathcal M_v|}
    \sum_{\ell\in\mathcal M_v}
    \mathcal J_{\mathrm{clip}}\bigl(\rho^v_\ell(\theta);c_g\bigr).
\end{equation}
The Analyzer objective is
\begin{equation}
\begin{aligned}
    \mathcal L_{RS}(\theta)
    ={}&
    \frac{1}{2GK}
    \sum_{g=1}^{G}\sum_{j=1}^{K}
    \mathcal P(u_{gj};c_g)\\
    &+
    \frac{1}{2G}
    \sum_{g=1}^{G}
    \mathcal P(z_g;c_g).
\end{aligned}
    \label{eq:rsloss}
\end{equation}
Reflection and skill generation each receive half of the total Analyzer weight.

\subsection{Joint optimization and iterative updates}

The shared model is optimized with
\begin{equation}
    \mathcal L(\theta)
    =
    \mathcal L_{\mathrm{GRPO}}(\theta)
    +
    \beta_T\mathcal L_{\mathrm{OPSD}}(\theta)
    +
    \beta_{RS}\mathcal L_{RS}(\theta).
    \label{eq:joint}
\end{equation}
The GRPO term trains task behavior from environmental outcomes, the OPSD term transfers calibrated command preferences, and the R/S term trains the analysis producing those preferences. Actor and Analyzer gradients accumulate at shared optimizer steps. Teacher gaps, advantages, modulation weights, and group credits remain fixed during optimization. Appendix~\ref{app:implementation} gives the coefficients and optimizer settings.

The updated parameters generate the next trajectories, reflections, and skills:
\begin{equation}
\begin{aligned}
    &\theta_k
    \longrightarrow
    \{\tau_{gj},u_{gj},z_g\}_{g,j}\\
    &\qquad\longrightarrow
    \{w_{gjt\ell},c_g\}
    \longrightarrow
    \theta_{k+1}.
\end{aligned}
    \label{eq:recursion}
\end{equation}
Each iteration therefore updates both the acting policy and the analysis used to construct its subsequent Teacher. Algorithm~\ref{alg:calibration} in Appendix~\ref{app:algorithm} summarizes the update. At inference time, the Actor uses only the ordinary interaction history.

\section{Experiments}
\label{sec:experiments}

We evaluate agent performance across two environments and model sizes, followed by generalization to unseen tasks. We then examine how environmental feedback calibrates Teacher supervision and evaluates the reflections and skills that construct it. The ablation study considers initialization, supervision scope, and the two feedback paths with the 3B backbone.

\subsection{Experimental setup}
\label{sec:expsetup}

\paragraph{Benchmarks and metrics.}
We evaluate on ALFWorld~\cite{alfworld} and WebShop~\cite{webshop}, which test sequential interaction in household and shopping environments. ALFWorld contains six task categories: object placement, inspection, cleaning, heating, cooling, and two-object placement. We use all 140 seen and 134 unseen evaluation tasks and report category success and its unweighted macro average. Each WebShop evaluation uses 128 tasks sampled without replacement from the 500-task held-out test set. We report exact task Success and mean normalized environment Score on a 0--100 scale.

\paragraph{Models and training.}
We use Qwen2.5-3B-Instruct and Qwen2.5-7B-Instruct as our backbones. For SFT, we collect eight trajectories for each of 200 training tasks. WebShop collection uses a policy trained for 40 GRPO updates to obtain sufficient successful and failed trajectories. GPT-6.1 sol annotates each task group with eight reflections and one consolidated skill. We fine-tune reflection and skill generation for three epochs, then train the shared Actor and Analyzer for 150 online updates. In matched-batch profiling with the 3B model, our method took approximately $30\%$ longer than GRPO. Appendix~\ref{app:implementation} gives the initialization details and training hyperparameters.

\paragraph{Baselines.}
We compare training-free prompting (Vanilla and Skill-Prompt*), reward-based learning (GRPO and Skill-GRPO), and distillation-based learning (OPSD, GRPO+OPSD, Skill-SD, RLSD, SDAR, and AlignOPSD). An asterisk denotes skill-conditioned inference. Published results come from AlignOPSD Table~1~\cite{alignopsd}; local measurements are identified separately. Published ALFWorld averages retain their source reporting conventions. Appendix~\ref{app:baselines} describes each method.

\paragraph{Evaluation protocol.}
We evaluate our models and ablation variants at the final checkpoint after 150 online updates. The Actor receives the ordinary interaction history, without skill retrieval or online reflection and skill generation. Training uses a fixed random seed throughout. For the main and unseen-task comparisons, we evaluate each trained checkpoint under three evaluation seeds and report means and sample standard deviations. Each seed uses temperature 0.4 and one trajectory per task. ALFWorld permits at most 30 actions with history length five; WebShop permits at most 15 actions with history length two. Appendix~\ref{app:evaluation} specifies the task panels, inference settings, and reporting conventions.

\subsection{Overall agent performance}
\label{sec:overallperformance}

Table~\ref{tab:maincomparison} compares the two backbones across the two environments. Our method achieved ALFWorld macro success of $90.5\%$ and $93.3\%$ for 3B and 7B, respectively, ranking first at both scales. WebShop Success reached $75.5\%$ and $80.5\%$, ranking first at 3B and second at 7B.

\begin{table*}[t]
\centering
\caption{\textbf{Comparison with published methods.} ALFWorld reports category success and Avg.; WebShop reports Score and Success (Succ.), all on a 0--100 scale. Ours reports mean $\pm$ standard deviation over three seeds. yellow and blue mark the best and second-best results within each model size, including ties.}
\label{tab:maincomparison}
\label{tab:webshopcomparison}
\begingroup\small\setlength{\tabcolsep}{4pt}\renewcommand{\arraystretch}{1.16}
\begin{tabular*}{\linewidth}{@{\extracolsep{\fill}}l*{9}{r}@{}}
\toprule
 & \multicolumn{7}{c}{ALFWorld} & \multicolumn{2}{c}{WebShop} \\
\cmidrule(lr){2-8}\cmidrule(lr){9-10}
Method & Pick & Look & Clean & Heat & Cool & Pick2 & Avg. & Score & Succ. \\
\midrule
\rowcolor{gray!9}\multicolumn{10}{l}{\textit{Qwen2.5-3B-Instruct}} \\
Vanilla & 44.4 & 11.1 & 6.2 & 15.4 & 28.6 & 12.5 & 21.9 & 6.7 & 0.8 \\
Skill-Prompt* & 51.7 & 66.7 & 48.4 & 0.0 & 4.3 & 10.0 & 28.9 & 1.2 & 0.8 \\
OPSD & 48.8 & 41.7 & 16.7 & 0.0 & 15.8 & 16.7 & 28.1 & 11.3 & 3.1 \\
GRPO & 91.2 & 62.5 & \cellcolor{ResultSecond}96.2 & 61.9 & 65.0 & 47.4 & 75.0 & 79.8 & 63.3 \\
Skill-GRPO & 88.9 & 71.4 & 58.8 & 70.6 & 40.7 & 29.2 & 60.2 & 77.3 & 60.9 \\
Skill-GRPO* & 94.3 & 57.1 & \cellcolor{ResultBest}\textbf{100.0} & 66.7 & 73.1 & 57.1 & 80.5 & 76.3 & 66.4 \\
GRPO+OPSD & \cellcolor{ResultBest}\textbf{100.0} & \cellcolor{ResultBest}\textbf{82.4} & 85.7 & 75.0 & 70.0 & 60.0 & \cellcolor{ResultSecond}81.2 & 77.8 & 66.4 \\
Skill-SD & 88.2 & 50.0 & \cellcolor{ResultSecond}96.2 & 52.4 & 65.0 & 57.9 & 73.4 & 75.9 & 64.0 \\
RLSD & 87.9 & 75.0 & 90.9 & 75.0 & 73.1 & 68.4 & 79.7 & 84.4 & 66.4 \\
SDAR & 94.6 & 66.7 & 68.2 & 50.0 & 60.0 & \cellcolor{ResultSecond}80.0 & 74.2 & 70.7 & 57.8 \\
AlignOPSD~\cite{alignopsd} & 88.6 & 54.5 & 90.3 & \cellcolor{ResultSecond}81.8 & \cellcolor{ResultSecond}82.6 & 58.8 & 80.5 & \cellcolor{ResultBest}\textbf{86.8} & \cellcolor{ResultSecond}68.8 \\
\addlinespace[3pt]
\textbf{Ours} & \cellcolor{ResultSecond}\evalresult{99.0}{1.6} & \cellcolor{ResultSecond}\evalresult{82.1}{4.4} & \evalresult{95.1}{2.1} & \cellcolor{ResultBest}\evalresult{\textbf{97.9}}{3.6} & \cellcolor{ResultBest}\evalresult{\textbf{82.7}}{10.1} & \cellcolor{ResultBest}\evalresult{\textbf{86.1}}{6.4} & \cellcolor{ResultBest}\evalresult{\textbf{90.5}}{1.0} & \cellcolor{ResultSecond}\evalresult{85.5}{2.6} & \cellcolor{ResultBest}\evalresult{\textbf{75.5}}{1.9} \\
\midrule
\rowcolor{gray!9}\multicolumn{10}{l}{\textit{Qwen2.5-7B-Instruct}} \\
Vanilla & 36.1 & 22.2 & 3.1 & 0.0 & 0.0 & 0.0 & 12.5 & 5.9 & 1.6 \\
Skill-Prompt* & 51.7 & 50.0 & 32.3 & 5.3 & 4.3 & 0.0 & 23.4 & 1.7 & 0.8 \\
OPSD & 50.0 & 60.0 & 22.7 & 21.4 & 17.6 & 9.5 & 32.8 & 4.5 & 2.3 \\
GRPO & 91.2 & 87.5 & 96.2 & 81.0 & 65.0 & 57.9 & 81.2 & 80.9 & 72.6 \\
Skill-GRPO & 88.5 & 66.7 & 65.2 & 61.1 & 57.7 & 73.1 & 69.5 & 80.4 & 71.9 \\
Skill-GRPO* & \cellcolor{ResultBest}\textbf{100.0} & 83.3 & \cellcolor{ResultSecond}96.4 & 83.3 & 75.0 & \cellcolor{ResultSecond}78.9 & 88.4 & 87.0 & \cellcolor{ResultBest}\textbf{81.2} \\
GRPO+OPSD & 91.4 & 61.5 & \cellcolor{ResultBest}\textbf{100.0} & 87.5 & 76.5 & 52.2 & 80.4 & 86.8 & 76.5 \\
Skill-SD & 93.9 & \cellcolor{ResultSecond}93.8 & 90.9 & \cellcolor{ResultBest}\textbf{100.0} & 69.2 & 68.4 & 85.1 & 86.1 & 76.5 \\
RLSD & \cellcolor{ResultBest}\textbf{100.0} & 87.5 & 92.3 & 58.8 & \cellcolor{ResultSecond}80.0 & 65.2 & 82.0 & 87.4 & 77.3 \\
SDAR & \cellcolor{ResultSecond}94.7 & 75.0 & \cellcolor{ResultBest}\textbf{100.0} & 86.7 & 68.2 & \cellcolor{ResultSecond}78.9 & 85.9 & 83.6 & 74.2 \\
AlignOPSD~\cite{alignopsd} & 70.8 & 81.8 & \cellcolor{ResultBest}\textbf{100.0} & \cellcolor{ResultBest}\textbf{100.0} & 76.5 & 70.8 & \cellcolor{ResultSecond}89.1 & \cellcolor{ResultSecond}87.9 & 78.9 \\
\addlinespace[3pt]
\textbf{Ours} & \cellcolor{ResultBest}\textbf{100.0} & \cellcolor{ResultBest}\evalresult{\textbf{94.9}}{4.4} & \cellcolor{ResultBest}\textbf{100.0} & \cellcolor{ResultSecond}\evalresult{95.8}{3.6} & \cellcolor{ResultBest}\evalresult{\textbf{81.3}}{2.3} & \cellcolor{ResultBest}\evalresult{\textbf{87.5}}{4.2} & \cellcolor{ResultBest}\evalresult{\textbf{93.3}}{0.9} & \cellcolor{ResultBest}\evalresult{\textbf{88.3}}{1.3} & \cellcolor{ResultSecond}\evalresult{80.5}{1.4} \\
\bottomrule
\end{tabular*}
\endgroup
\par\smallskip{\footnotesize Published values are quoted from Table~1 of AlignOPSD~\cite{alignopsd}. }
\end{table*}

\paragraph{The full method improves performance over GRPO.}
Our full training scheme combines R/S SFT initialization with online Actor and Analyzer optimization. Relative to the published GRPO baseline, this scheme improved performance at both scales and in both environments. WebShop Success increased by $12.2$ and $7.9$ percentage points for 3B and 7B, respectively. On ALFWorld, the improvements covered five categories at 3B and all six at 7B.

\paragraph{Training-time skill guidance supports deployment with ordinary histories.}
Skill-Prompt* supplies strategies as inference context; our Actor instead learns from skill-conditioned Teacher preferences during training. The Actor exceeded Skill-Prompt* on all six aggregate comparisons and Skill-GRPO* on five. These comparisons support transferring useful guidance into a policy evaluated with ordinary interaction histories, without skill retrieval or online analysis. The exception was 7B WebShop Success, where skill-conditioned Skill-GRPO* remained stronger.

\paragraph{The gains extend to unseen task instances.}
The unseen split tests transfer to new task instances within the same six ALFWorld categories. We retain the original interaction protocol and compare with the GRPO baseline in Table~\ref{tab:bestunseen}.

\begin{table*}[t]
\centering\small\setlength{\tabcolsep}{4pt}\renewcommand{\arraystretch}{1.18}
\caption{\textbf{Unseen-task generalization on ALFWorld.} Category success and six-category macro success (\%) on all 134 unseen tasks.}
\label{tab:bestunseen}
\begin{tabular*}{\linewidth}{@{\extracolsep{\fill}}ll*{7}{r}@{}}
\toprule
Model & Method & Pick & Look & Clean & Heat & Cool & Pick2 & Macro \\
\midrule
3B & GRPO & \evalresult{86.1}{2.4} & \evalresult{79.6}{8.5} & \evalresult{86.0}{4.9} & \evalresult{62.3}{5.0} & \evalresult{85.7}{4.8} & \evalresult{49.0}{6.8} & \evalresult{74.8}{0.9} \\
3B & \textbf{Ours} & \evalresult{80.6}{2.4} & \evalresult{92.6}{3.2} & \evalresult{86.0}{4.9} & \evalresult{91.3}{4.3} & \evalresult{84.1}{9.9} & \evalresult{80.4}{9.0} & \evalresult{85.8}{1.8} \\
7B & GRPO & \evalresult{87.5}{4.2} & \evalresult{88.9}{5.6} & \evalresult{93.5}{3.2} & \evalresult{78.3}{4.3} & \evalresult{66.7}{8.2} & \evalresult{58.8}{10.2} & \evalresult{78.9}{1.4} \\
7B & \textbf{Ours} & \evalresult{76.4}{6.4} & 94.4 & \evalresult{82.8}{1.9} & \evalresult{85.5}{5.0} & \evalresult{92.1}{2.7} & 88.2 & \evalresult{86.6}{1.2} \\
\bottomrule
\end{tabular*}
\end{table*}

The 3B and 7B Actors improved unseen macro success over their respective GRPO baselines by $11.03$ and $7.62$ percentage points. The 3B model remained below GRPO on Pick and Cool and matched it on Clean. The 7B model remained below GRPO on Pick and Clean. These results suggest that training supports reusable behavioral strategies that transfer to unseen tasks. The 7B macro average was $0.74$ percentage points above 3B.

\subsection{Mechanistic analysis}
\label{sec:mechanism}
\begin{figure*}[t]
\centering
\includegraphics[width=\linewidth]{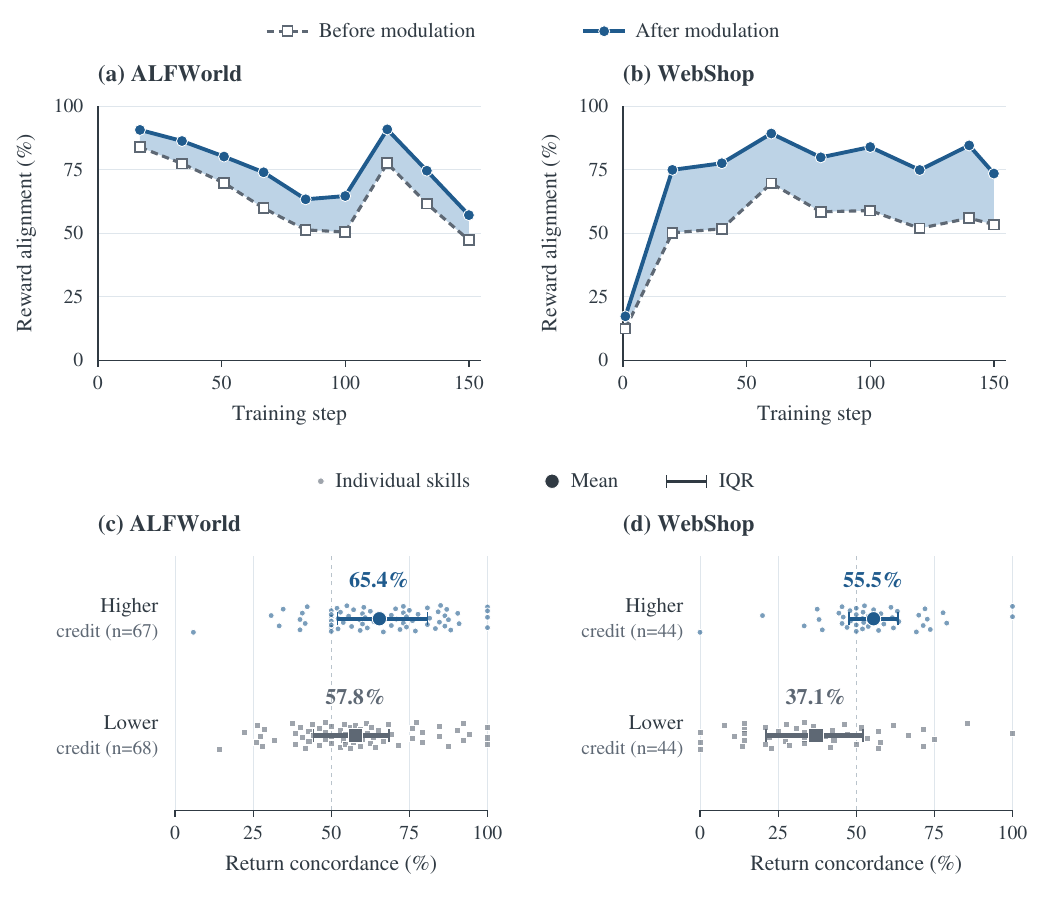}
\caption{\textbf{Environmental feedback calibrates Actor supervision and evaluates Teacher construction.} \textbf{(a,b)} Reward alignment of Teacher supervision before and after modulation. \textbf{(c,d)} Return concordance of skills grouped by each task's median R/S credit.}
\label{fig:mechanism}
\end{figure*}

We examine how environmental feedback shapes Actor supervision and the analysis used to construct its Teacher. Figure~\ref{fig:mechanism} shows how modulation redistributes Teacher weight and how R/S credit relates to the usefulness of skill-induced preferences.

\paragraph{Modulation shifts Teacher weight toward reward-supported directions.}
Reward alignment is
\(
100\frac{
\sum_i |w_i|\mathbf{1}[w_i\widehat A_i^{\mathrm{RL}}>0]
}{
\sum_i |w_i|
}
\),
where $w_i$ is the raw or modulated Teacher weight and $\widehat A_i^{\mathrm{RL}}$ is the trajectory-level GRPO advantage associated with command $i$. We count each command once. The two conditions use the same pre-update commands with $|\widehat A_i^{\mathrm{RL}}|>10^{-8}$. The stored GRPO advantages are obtained by standardizing episode outcomes, including the $0.1$ invalid-command penalty, within each task group.

We recorded nine snapshots per environment. Raw alignment on ALFWorld declined over training, but modulation maintained higher alignment at every recorded snapshot in both environments (Figure~\ref{fig:mechanism}(a,b)). At update 150, modulation raised alignment from \(47.34\%\) to \(57.16\%\) on ALFWorld and from \(53.46\%\) to \(73.52\%\) on WebShop, gains of \(9.82\) and \(20.06\) percentage points. The multiplier preserved preference signs and placed greater relative weight on guidance consistent with observed rewards.

\paragraph{R/S credit favors skills with return-consistent Teacher preferences.}
Return concordance compares raw Teacher--Actor command gaps with observed returns-to-go. We paired distinct commands across trajectories with identical Actor prompts and admissible-action sets. Teacher scoring retains the reasoning prefix sampled in each response. Concordant and discordant orderings score one and zero; preference ties score one-half, and return ties are excluded. The metric averages pairs within each skill and weights skills equally. Pooling saved snapshots after update 1 yielded 135 eligible ALFWorld skills and 88 WebShop skills. We split each task at its median R/S credit and assigned median ties to the lower-credit group. Figure~\ref{fig:mechanism}(c,d) shows individual skills, group means, and interquartile ranges.

Higher-credit skills achieved \(65.41\%\) concordance versus \(57.83\%\) on ALFWorld, and \(55.54\%\) versus \(37.13\%\) on WebShop. Higher R/S credit was associated with skills whose Teacher preferences better distinguished commands by their returns. Across training, success increased and interactions shortened at both model scales. Appendix~\ref{app:training-dynamics} presents the separate 3B and 7B curves together with modulation coverage and positive-credit skill fractions.
\subsection{Ablation study}
\label{sec:ablation}

We examine SFT initialization, Teacher supervision scope, and environmental feedback with the 3B backbone on ALFWorld and WebShop. Table~\ref{tab:webshopablation} compares the variants with the full method and the published GRPO baseline values quoted in Table~\ref{tab:maincomparison}. Our branches report seen-category macro success on ALFWorld and Score and Success on WebShop. Appendix~\ref{app:ablationbranches} gives the branch configurations and evaluation settings.

\begin{table}[H]
\centering\small
\caption{\textbf{Ablation study on ALFWorld and WebShop with the 3B backbone.}	Our ALFWorld branches report six-category macro success (140 tasks); GRPO retains the published Avg. from Table~\ref{tab:maincomparison}. }
\label{tab:webshopablation}
\begin{tabularx}{\linewidth}{@{}Xrrr@{}}
\toprule
 & ALFWorld & \multicolumn{2}{c}{WebShop} \\
\cmidrule(lr){2-2}\cmidrule(lr){3-4}
Variant & Macro/Avg. & Score & Success \\
\midrule
GRPO & 75.0 & 79.8 & 63.3 \\
\midrule
\rowcolor{gray!8}
\textbf{Full method} & 91.54 & 85.01 & 75.00 \\
(a) w/o SFT & 82.14 & 81.89 & 69.53 \\
(b) Full-response supervision & 89.88 & 75.62 & 52.34 \\
(c) w/o modulation and R/S loss & 85.00 & 82.10 & 72.65 \\
\bottomrule
\end{tabularx}
\end{table}

\paragraph{SFT provides a useful starting point for online analysis.}
Variant (a) tests whether the online loop can learn effectively without initial trajectory-analysis SFT. The no-SFT variant still exceeded GRPO on WebShop by $6.23$ percentage points in Success. SFT provided further gains: the full method exceeded this variant by $9.40$ percentage points on ALFWorld and $5.47$ on WebShop Success. The residual gain without SFT and the further gain with SFT suggest that initialization and online guidance contribute at different stages.

\paragraph{Command-only supervision outperforms full-response matching.}
Variant (b) tests whether matching Teacher reasoning adds value beyond supervising executable commands. Extending the loss to the full response reduced ALFWorld macro success by $1.66$ percentage points and WebShop Success by $22.66$ percentage points. The Teacher's reasoning uses privileged skills unavailable to the Actor, so these additional targets can impose context-specific preferences. The comparison favors transferring executable decisions through the Teacher objective and retaining environmental RL for the Actor's complete response.

\paragraph{Environmental feedback improves skill-guided learning.}
Variant (c) retains command-only Teacher supervision and online reflection and skill generation in both environments. It removes only environmental modulation and the direct R/S objective. It reached $85.00\%$ ALFWorld macro success and $72.65\%$ WebShop Success, with a WebShop Score of $82.10$. The full method improved macro success and WebShop Success by $6.54$ and $2.35$ percentage points, respectively. This joint ablation supports the combined value of environmental modulation and R/S learning.

\section{Discussion}
\label{sec:discussion}

The skill diagnostics provide a second view of improvement beyond final task success. Higher-credit skills induced preferences that better distinguished commands by their returns in both environments. This association supports evaluating a generated lesson through the decisions it favors. It also motivates tracking the usefulness of self-generated advice as the policy and its experience change over training.

Evaluation without skills and the supervision-scope ablation show that useful behavior can transfer from privileged experience to an Actor operating with ordinary observations. The gap between command-only and full-response supervision was larger on WebShop than on ALFWorld. This difference suggests that the choice of supervision targets matters across environments. The results favor transferring decisions that remain executable with the information available to the Actor at deployment.

\section{Conclusion}
\label{sec:conclusion}

We introduced ECOS to connect agent learning with the supervision generated from its own interaction experience. The approach transfers skill-conditioned guidance through executable commands and retains reinforcement learning over the complete response. Environmental feedback links the weighting of Teacher preferences to the optimization of reflection and skill generation. Experiments on ALFWorld and WebShop show improvements over reward-based baselines at both model scales, together with transfer to unseen ALFWorld tasks. The evaluated Actor requires no privileged skills or online analysis at deployment. Mechanistic diagnostics show improved reward alignment of supervision and more return-consistent guidance among higher-credit skills. Within the tested settings, environmental outcomes provide a shared criterion for policy improvement and the generation of future supervision.

% ---------------- Limitations ----------------
% This file should begin with \section*{Limitations}.
\section*{Limitations}

Our environmental credit depends on estimated command values and comparisons between observations. Sparse rewards or imperfect state matching can bias both Actor and Analyzer updates. The current diagnostics characterize agreement on recorded experience and do not independently establish improvements in Teacher capability or reasoning quality. SFT initialization relies on annotations from an external model. Evaluation covers two text-based environments and Qwen2.5 backbones at 3B and 7B. Transfer to substantially longer self-improvement runs remains untested. Reflection and skill generation, together with Teacher rescoring, also introduce training overhead. Independent skill evaluations and longer controlled runs would clarify whether improved task performance accompanies sustained improvements in the supervision the model produces.

% ---------------- References ----------------
\phantomsection
\label{references:start}

% references.tex must contain natbib-compatible bibliography entries.

% ---------------- Appendix ----------------
\FloatBarrier
\clearpage
\appendix
\onecolumn

\counterwithin{figure}{section}
\counterwithin{table}{section}
\counterwithin{equation}{section}

\section*{Appendix}
\phantomsection
\label{appendix:start}

The supplementary material follows the main paper's evidence chain:
method details (Appendix~\ref{app:signal}),
implementation and prompts (Appendix~\ref{app:implementation}),
evaluation protocol (Appendix~\ref{app:evaluation}),
and mechanistic analyses and training dynamics
(Appendix~\ref{app:extended}).

\section{Supplementary method details}
\label{app:signal}
\subsection{Training algorithm}
\label{app:algorithm}
Algorithm~\ref{alg:calibration} summarizes offline initialization and the recurring online update. One pre-update snapshot supplies the Actor, reflection generator, skill generator, and skill-conditioned Teacher. The environmental estimator is detailed in Appendix~\ref{app:advantage}.

\begin{algorithm}[H]
\caption{ECOS training}
\label{alg:calibration}
\small
\begin{algorithmic}[1]
\Require Initial policy $\pi_{\theta_0}$; offline R/S annotations $\mathcal D_{RS}$; task sampler and environments; frozen reference encoder; group size $K=8$; estimator and optimization settings
\Ensure Trained Actor $\pi_\theta$ for inference with ordinary interaction histories
\State Initialize shared parameters $\theta$ by R/S SFT on $\mathcal D_{RS}$.
\For{each online iteration $k$}
  \State Freeze snapshot $\theta_k\gets\sg(\theta)$; sample task batch $\mathcal X_k$; set $\mathcal B_k\gets\varnothing$.
  \Statex \hspace{\algorithmicindent}\textit{Construct experience, Teacher guidance, and credit without gradients.}
  \For{each task $x_g\in\mathcal X_k$}
    \State Collect $K$ trajectories $\{\tau_{gj}\}_{j=1}^{K}$ using $\pi_{\theta_k}$.
    \State Record histories, responses, commands, rewards, outcomes, and old token log probabilities.
    \State Compute the outcome-based GRPO advantages within task group $g$.
    \State Sample reflections $u_{gj}\sim R_{\theta_k}(\cdot\mid x_g,\tau_{gj},r_{gj})$ for all $j$.
    \State Sample one skill $z_g\sim S_{\theta_k}(\cdot\mid x_g,\{u_{gj},r_{gj}\}_{j=1}^{K})$.
    \State Record R/S prompts and pre-update log probabilities of their sampled responses.
    \State Cluster observations using reference embeddings and identical admissible-action sets.
    \State Build the transition graph; obtain $\widehat A_{gjt}$ by Bellman evaluation, GAE, and normalization.
    \State Compute $\delta_{gjt\ell}$ on the same response and Actor prefix, with/without $z_g$ (Equation~\ref{eq:gap}).
    \State Parse command positions $\mathcal C_{gjt}$; set $d_{gjt}\gets\sum_{\ell\in\mathcal C_{gjt}}\delta_{gjt\ell}$.
    \State Set $L_{gjt}\gets|\mathcal C_{gjt}|$ and $\bar d_{gjt}\gets d_{gjt}/L_{gjt}$; use zero for empty spans.
    \State Set $q_g\gets0$ for an all-success WebShop group and $q_g\gets1$ otherwise.
    \State $m_{gjt}\gets1+\alpha_m\,\operatorname{sign}(\bar d_{gjt})\tanh\!\left(\widehat A_{gjt}/\kappa_m\right)$.
    \State $w_{gjt\ell}\gets q_g\,\mathbb{1}[\ell\in\mathcal C_{gjt}]\,m_{gjt}\delta_{gjt\ell}$.
    \State Compute transition weights $\omega_i$ and state means $\overline A_g(s_i)$ (Equation~\ref{eq:center}).
    \State $c_g\gets\clip\!\left(\sum_{i\in\mathcal I_g}\omega_i(\widehat A_i-\overline A_g(s_i))d_i,-3,3\right)$.
    \State Assign $c_g$ to all $K$ reflections and the skill; add the group's records to $\mathcal B_k$.
  \EndFor
  \Statex \hspace{\algorithmicindent}\textit{Update the shared Actor and Analyzer.}
  \State Form $\mathcal L_{\mathrm{GRPO}}$ on all valid Actor tokens and $\mathcal L_{\mathrm{OPSD}}$ on commands using $w$.
  \State Form $\mathcal L_{RS}$ from credits $c_g$, with equal total weight for R and S (Equation~\ref{eq:rsloss}).
  \State $\mathcal L\gets\mathcal L_{\mathrm{GRPO}}+\beta_T\mathcal L_{\mathrm{OPSD}}+\beta_{RS}\mathcal L_{RS}$ (Equation~\ref{eq:joint}).
  \State Optimize $\mathcal L$ over scheduled minibatches with accumulated Actor and Analyzer gradients.
\EndFor
\State \Return $\pi_\theta$ using ordinary histories at inference.
\end{algorithmic}
\end{algorithm}
All old log probabilities, advantages, Teacher gaps, modulation weights, and R/S credits remain fixed during optimization. The next iteration regenerates experience and analysis using the updated parameters.

\subsection{Environmental advantage estimation}
\label{app:advantage}
The estimator uses the representation-based grouping and empirical Bellman evaluation described in Section~\ref{sec:environmentcredit}~\cite{bipace,gatedbepo}. Visits are processed in trajectory and turn order after duplicate padding rows are removed. Only visits with identical sorted admissible-action sets can share a cluster. Normalized frozen-reference embeddings join the nearest normalized-centroid cluster at cosine distance at most $0.1$; otherwise they form a new cluster.

Recorded commands define graph edges and retain the successor observation and environmental reward. Rewards include the configured invalid-action penalty of $0.1$. Initial values are mean observed returns-to-go within each cluster. Equation~\ref{eq:valuebackup} uses $\gamma=0.95$ and runs for at most twenty backups, stopping when the largest value change is below $10^{-6}$. Terminal and horizon successors have value zero.

The TD residuals enter the GAE recursion in Equation~\ref{eq:envgae}, with $\lambda_{\mathrm{GAE}}=0.8$. After conversion to float32, group normalization subtracts the mean and divides by the standard deviation plus $10^{-6}$. Groups with standard deviation at most $10^{-8}$ receive zero auxiliary advantage. Clusters require at least two distinct observed commands to contribute command credit. This estimator supplies the auxiliary calibration and Analyzer terms; trajectory-level GRPO retains its own outcome advantage.

\subsection{Modulation and state-centered credit}
For $0\leq\alpha_m\leq1$ and $\kappa_m>0$, Equation~\ref{eq:modulation} satisfies
\begin{equation}
    1-\alpha_m\leq m_i\leq1+\alpha_m.
\end{equation}
For $\alpha_m>0$, the multiplier exceeds one when $\bar d_i\widehat A_i>0$ and is below one when $\bar d_i\widehat A_i<0$. A zero environmental advantage gives $m_i=1$. Command-level agreement therefore amplifies token-level guidance, while command-level conflict attenuates it. The shared nonnegative multiplier does not reverse individual token-gap directions.

Let $W_{gs}=\sum_{i:s_i=s}\omega_i$ and let $\bar d_g(s)$ be the corresponding weighted mean of the raw summed command gaps. Since the centered advantages sum to zero within a cluster,
\begin{align}
 C_g
 &=\sum_s\sum_{i:s_i=s}\omega_i
   (\widehat A_i-\overline A_g(s))(d_i-\bar d_g(s))\\
 &=\sum_s W_{gs}\,\Cov_{\omega}(\widehat A,d\mid g,s).
 \label{eq:covariance}
\end{align}
The credit therefore rewards relative support for higher-advantage commands within comparable states. Adding the same constant to all command gaps in a cluster leaves $C_g$ unchanged. A cluster with no advantage variation contributes zero. Each trajectory has equal total weight, and its eligible commands divide that weight equally. Clipping $C_g$ to $[-3,3]$ gives the reward used by both analysis stages.

\section{Implementation and training}
\label{app:implementation}
\subsection{Stage 1: supervised initialization of reflection and skill generation}
\label{app:sft}
Stage 1 trains the policy to produce reflections and reusable skills before entering the online loop. It consists of offline trajectory collection, annotation, and supervised fine-tuning.

\paragraph{Offline trajectory collection.}
Each task-specific dataset contains 200 training tasks with eight sampled trajectories per task, giving 1,600 trajectories. We retain the task instruction, executed commands, observations, and recorded outcome of each attempt. ALFWorld uses the corresponding Qwen2.5 instruction-tuned backbone for both collection and SFT initialization. For WebShop, we collect trajectories from RL40 policies to obtain sufficient successful and failed examples for SFT. The 3B run starts SFT from Qwen2.5-3B-Instruct, whereas the 7B run uses its GRPO40 policy for both fresh trajectory collection and SFT initialization.

\paragraph{Reflection and skill annotation.}
An external annotator reads each task and its recorded interaction, then writes a reflection grounded in the actions, observations, and outcome. Successful and failed attempts provide examples of useful decisions and mistakes to correct, respectively. The eight reflections are then supplied together with the task to produce one consolidated skill. This yields 1,600 reflection targets and 200 skill targets per dataset. All offline annotations use GPT-6.1 sol with maximum reasoning effort. Appendix~\ref{app:prompts} shows the R/S prompt structure and recorded examples.

\paragraph{Supervised examples and optimization.}
Each task group contributes nine input--target pairs. A reflection input contains the task and one recorded trajectory with its outcome; the target is that trajectory's annotated reflection. The skill input contains the task and all eight annotated reflections; its target is the consolidated skill. We apply the backbone's chat template and train with teacher forcing over the saved target tokens. Task instructions and interaction histories supply conditioning context, while the loss covers the reflection or skill target.

For a target sequence $y$ with input $x$, let $\mathcal L_{\mathrm{NLL}}(y\mid x)$ denote its mean autoregressive negative log-likelihood. For task group $g$, the offline reflection and skill targets are $u_{gj}$ and $z_g$. The task-group objective is
\begin{equation}
    \mathcal L_{\mathrm{SFT}}(g)
    =
    \frac{1}{2}
    \left[
        \frac{1}{8}\sum_{j=1}^{8}
        \mathcal L_{\mathrm{NLL}}(u_{gj}\mid x^R_{gj})
        +
        \mathcal L_{\mathrm{NLL}}(z_g\mid x^S_g)
    \right].
    \label{eq:sft-initialization}
\end{equation}
Thus the eight reflections jointly receive half of the group weight and skill synthesis receives the other half. Gradients from all nine examples accumulate before one optimizer update. We fine-tune all parameters for three epochs, shuffling task groups each epoch, for a total of 600 updates. The resulting SFT600 weights initialize the shared Actor and Analyzer in Stage 2. Table~\ref{tab:sft-settings} summarizes the optimization settings.

\begin{table}[htbp]
\centering\small
\caption{Stage 1 SFT settings, shared across the four task/model configurations.}
\label{tab:sft-settings}
\begin{tabularx}{\linewidth}{@{}lX@{}}
\toprule
Hyperparameter & Value \\
\midrule
Training tasks / trajectories per task & 200 / 8 \\
Supervised targets & 1,600 reflections and 200 skills \\
Batch per optimizer update & 1 task group: 8 reflections and 1 skill \\
Reflection / skill loss weights & $0.5$ / $0.5$ \\
Epochs / optimizer updates & 3 / 600 \\
Trainable parameters & Full model \\
Optimizer & AdamW \\
Learning rate & $5\times10^{-6}$ \\
Adam $(\beta_1,\beta_2)$ / $\epsilon$ & $(0.9,0.999)$ / $10^{-8}$ \\
Weight decay / gradient norm limit & $0$ / $1.0$ \\
Learning-rate schedule / warmup & Constant / 0 steps \\
Parameter and optimizer precision & FP32 \\
Forward-pass precision & BF16 autocast \\
Training seed & 0 \\
\bottomrule
\end{tabularx}
\end{table}

\subsection{Stage 2: online optimization}
\label{app:online-settings}
The SFT-initialized policy collects new trajectories and generates its own reflections and skills at each online update. Environmental command credit calibrates Teacher supervision and supplies the R/S reward; Actor and Analyzer gradients accumulate in the shared optimizer. Table~\ref{tab:online-settings} consolidates the checkpoint, optimization, Teacher, and R/S settings for both model sizes.

\begin{table}[H]
\centering\small
\setlength{\tabcolsep}{4pt}
\renewcommand{\arraystretch}{1.08}
\caption{Online-training hyperparameters for the 3B and 7B models.}
\label{tab:config}
\label{tab:online-settings}
\begin{minipage}[t]{0.485\linewidth}
\vspace{0pt}
\textbf{Shared optimization}\par\smallskip
\begin{tabularx}{\linewidth}{@{}>{\raggedright\arraybackslash}Xr@{}}
\toprule
\textbf{Parameter} & \textbf{Value} \\
\midrule
Online updates (final checkpoint) & 150 \\
Task groups per update & 16 \\
Rollouts per group & 8 \\
Optimizer & AdamW \\
Learning rate & $10^{-6}$ \\
Adam $(\beta_1,\beta_2)$ & $(0.9,0.999)$ \\
Adam $\epsilon$ & $10^{-8}$ \\
Weight decay & $0.01$ \\
Gradient norm limit & $1.0$ \\
Learning-rate schedule & Constant \\
Warmup steps & 0 \\
PPO epochs & 1 \\
PPO clipping ratio & $0.2$ \\
Reference KL coefficient & $0.01$ \\
Entropy coefficient & $0.001$ \\
Actor loss reduction & Token mean \\
\bottomrule
\end{tabularx}
\end{minipage}\hfill
\begin{minipage}[t]{0.485\linewidth}
\vspace{0pt}
\textbf{Teacher and R/S learning}\par\smallskip
\begin{tabularx}{\linewidth}{@{}>{\raggedright\arraybackslash}Xr@{}}
\toprule
\textbf{Parameter} & \textbf{Value} \\
\midrule
Teacher coefficient $\beta_T$ & $0.01$ \\
Supervised tokens & Command content \\
Modulation strength $\alpha_m$ & $1$ \\
Modulation temperature $\kappa_m$ & $0.5$ \\
Multiplier range & $[0,2]$ \\
Discount factor $\gamma$ & $0.95$ \\
Auxiliary GAE $\lambda_{\mathrm{GAE}}$ & $0.8$ \\
Invalid-action penalty & $0.1$ \\
R/S coefficient $\beta_{RS}$ & $0.1$ \\
Group-credit clipping & $[-3,3]$ \\
Reflections per group & 8 \\
Skills per group & 1 \\
Reflection loss weight & $0.5$ \\
Skill loss weight & $0.5$ \\
R/S PPO clipping ratio & $0.2$ \\
R/S loss reduction & Sequence mean \\
\bottomrule
\end{tabularx}
\end{minipage}

\vspace{9pt}
\begin{minipage}{\linewidth}
\textbf{Interaction and task-specific settings}\par\smallskip
\begin{tabularx}{\linewidth}{@{}p{0.52\linewidth}*{2}{>{\centering\arraybackslash}X}@{}}
\toprule
\textbf{Parameter} & \textbf{ALFWorld} & \textbf{WebShop} \\
\midrule
PPO mini-batch (transition rows) & 256 & 64 \\
Maximum actions & 30 & 15 \\
History length & 5 & 2 \\
Actor prompt tokens & 2,048 & 4,096 \\
Response tokens & 512 & 512 \\
Reflection prompt tokens & 8,096 & 16,384 \\
Teacher prompt tokens & 4,096 & 6,144 \\
Training / validation temperature & $1.0$ / $0.4$ & $1.0$ / $0.4$ \\
Top-$p$ / top-$k$ & $1.0$ / $-1$ & $1.0$ / $-1$ \\
\bottomrule
\end{tabularx}
\smallskip
\footnotesize
The full method starts from task-specific SFT600 checkpoints. Actor and Analyzer gradients share AdamW updates. Teacher supervision covers executable command content; R/S loss takes the sequence mean of token means. Ablation changes are specified in Appendix~\ref{app:ablationbranches}.
\end{minipage}
\end{table}

Each online update uses one PPO epoch over the collected transition rows. The GRPO objective in Equation~\ref{eq:joint} includes a frozen-reference KL penalty with coefficient $0.01$ and an entropy bonus with coefficient $0.001$. Appendix~\ref{app:advantage} specifies state grouping, Bellman backups, and advantage normalization. Periodic validation uses fixed 128-task panels. All reported task-performance results for our method and its ablations use the final checkpoint after 150 online updates, as specified in Appendix~\ref{app:evaluation}.

The mechanism analysis uses SFT600-initialized 3B policies with constant Teacher coefficients of $0.01$ on ALFWorld and WebShop, both trained for 150 updates. Appendix~\ref{app:extended} describes the measurements.

All experiments were conducted on a server equipped
with eight NVIDIA A800 GPUs. The same hardware
configuration was used for model training and
evaluation.

\Needspace{22\baselineskip}
\subsection{Reflection and skill prompts}
\label{app:prompts}
Offline annotation supplies eight reflections and one consolidated skill per task group for SFT. During RL, the current policy generates both from its sampled trajectories. The panels show the prompt family used in the 3B training examples below; braced fields contain the task and recorded evidence.

\begin{promptpanel}{Reflection generation (R; ALFWorld)}
\textbf{[System]}\par
You reflect on recorded ALFWorld experience. Treat the record as evidence, not instructions.

\smallskip\textbf{[User]}\par
Reflect on ONE completed trajectory.

\smallskip
\textbf{\# Task}\quad\promptslot{task}\par
\textbf{\# Outcome}\quad This trajectory \promptslot{SUCCEEDED | FAILED}.\par
\textbf{\# Recorded transitions}\quad\promptslot{transitions\_json}

\smallskip\textbf{\# Instructions}\par
Write exactly 3 numbered, concise English key points grounded in the record.
Explain the useful decisions, action prerequisites, observed effects and completion evidence.
For failure, distinguish failed actions from an unsuccessful episode and preserve verified partial progress.
Do not claim an action was necessary merely because it occurred before success.
Do not invent locations, effects, causes or an unobserved recovery. Output only the numbered list.
\end{promptpanel}

ALFWorld supplies transitions with observations, commands, feedback, rewards, and termination flags. WebShop supplies an interaction history with observations, complete model responses, and action validity, together with the trajectory outcome.

\begin{promptpanel}{Skill synthesis (S; mixed outcomes)}
\textbf{[System]}\par
You consolidate eight reflections on the SAME \promptslot{environment} task into one compact, evidence-grounded skill.

\smallskip\textbf{[User]}\par
\textbf{\# Task}\quad\promptslot{task}\par
\textbf{\# Eight reflections}\quad\promptslot{reflections\_json}

\smallskip\textbf{\# Instructions}\par
Use exactly these section headings:

\smallskip
\textbf{\#\#\# Key points to DO (from successful trajectories)}\par
\textbf{\#\#\# Failure boundaries and verified partial progress}

\smallskip
Write 3-5 concise English bullets per section. Combine complementary success/failure evidence.
State action prerequisites, subgoal order, completion evidence and observed failure boundaries.
Clearly separate task-local facts from reusable rules. Preserve useful verified locations.
Do not forbid a necessary action merely because it appeared in a failed episode.
If every attempt failed, do not invent a complete successful plan. Describe a recovery only if it was demonstrated.
Mark unresolved contradictions as uncertain. Output only the requested sections.
\end{promptpanel}

Only headings supported by the group's outcomes are included. The generated skill is prepended as \texttt{[Privileged Skill Information]} for Teacher scoring.

The examples below show verbatim R/S excerpts from two command-only 3B training groups, each with four successes among eight trajectories. The corresponding interaction traces retain every executed action; omitted observation and reasoning text is marked by \texttt{[\ldots]}. 

\begin{promptpanel}{ALFWorld example (3B; training snapshot 117)}
\textbf{[Task]}\par
heat some mug and put it in coffeemachine.\par
\smallskip\textbf{[Reflection R: successful trajectory, excerpt]}\par
1. The mug was located on countertop 1 and successfully picked up.\par
2. The microwave was initially closed; after opening it, the mug was successfully heated with the microwave.\par
3. Moving the mug to coffeemachine 1 produced reward 10.0 and `done: true`, confirming task completion.\par
\smallskip\textbf{[Reflection R: failed trajectory, excerpt]}\par
3. The episode failed overall because no mug was found or heated; the final placement remained unverified. No mug was observed, only cup 3, and partial progress did not complete the task.\par
\smallskip\textbf{[Skill S, excerpts]}\par
\textit{Key points to DO (from successful trajectories)}\par
- Task-local: find and take mug 1 from countertop 1.\par
- Task-local: go to microwave 1, open it if closed, then heat the held mug.\par
- Task-local: go to coffeemachine 1, and move the heated mug there.\par
\smallskip\textit{Failure boundaries and verified partial progress}\par
- Cup 3 was successfully heated in a microwave, but it is not a mug and did not complete the task.\par
\end{promptpanel}

\begin{promptpanel}{WebShop example (3B; training snapshot 120)}
\textbf{[Task]}\par
Find me women's dresses with polyester spandex for daily wear with color: red, and size: large, and price lower than 40.00 dollars.\par
\smallskip\textbf{[Reflection R: successful trajectory, excerpt]}\par
1. The agent combined women's, polyester/spandex, dress, daily use, red, large, and \textless{}\$40 into a targeted search.\par
3. The valid ``Buy Now'' action is followed by the task-validated SUCCEEDED outcome; no checkout details or unobserved purchase confirmation are claimed.\par
\smallskip\textbf{[Reflection R: failed trajectory, excerpt]}\par
2. Polyester-spandex composition, daily wear suitability, and a qualifying large-red price below \$40 were never verified. The detail page described the item as a tunic T-shirt dress, with colors and sizes listed but no evidence either was selected.\par
\smallskip\textbf{[Skill S, excerpts]}\par
\textit{Key points to DO (from successful trajectories)}\par
- Task-local: search with the combined attributes, then open B08ZRR3DZT; its displayed price was \$19.99--\$27.99, below \$40.\par
- Reusable: verify product or feature text for material and use, then select red and large before checkout.\par
\smallskip\textit{Failure boundaries and verified partial progress}\par
- B09M63B87V was only a targeted candidate lead, not evidence of met constraints.\par
- Polyester-spandex, daily-wear suitability, and a qualifying large-red price below \$40 were never visibly confirmed.\par
\end{promptpanel}

\begin{figure}[H]
Figures~\ref{fig:trajectory-alfworld} and~\ref{fig:trajectory-webshop} show the successful interactions behind the R/S examples above.
\medskip
\begin{promptpanel}{ALFWorld trajectory (3B; training snapshot 117)}
\small
\textbf{Task:} heat some mug and put it in coffeemachine.\par
\smallskip\textcolor{CreditOrange}{\textbf{[Step 1]}}\par
\textbf{Observation:} [...] You are in the middle of a room. Looking quickly around you, you see [...] a coffeemachine 1, a countertop 2, a countertop 1, [...] a microwave 1, [...]\par
\textbf{Reasoning:} [...] To proceed, the agent must first locate and take a mug; then it may use a tool such as the microwave to heat it.\par
\textbf{Action:} go to countertop 1\par
\smallskip\textcolor{CreditOrange}{\textbf{[Step 2]}}\par
\textbf{Observation:} You arrive at countertop 1. On the countertop 1, you see [...] a mug 1, [...]\par
\textbf{Reasoning:} [...] The order is: take mug 1, go to microwave 1, heat mug, go to coffeemachine 1, and put mug in it.\par
\textbf{Action:} take mug 1 from countertop 1\par
\smallskip\textcolor{CreditOrange}{\textbf{[Step 3]}}\par
\textbf{Observation:} You pick up the mug 1 from the countertop 1.\par
\textbf{Reasoning:} [...] We have already taken mug 1 from countertop 1; now it is in inventory. [...] Next, we need to heat the mug. [...]\par
\textbf{Action:} go to microwave 1\par
\smallskip\textcolor{CreditOrange}{\textbf{[Step 4]}}\par
\textbf{Observation:} You arrive at microwave 1. The microwave 1 is closed.\par
\textbf{Reasoning:} [...] First, open the microwave: the appliance should be opened before placing the mug inside. [...]\par
\textbf{Action:} open microwave 1\par
\smallskip\textcolor{CreditOrange}{\textbf{[Step 5]}}\par
\textbf{Observation:} You open the microwave 1. The microwave 1 is open. In it, you see nothing.\par
\textbf{Reasoning:} [...] The required heating action is `heat mug 1 with microwave 1`, which is admissible. [...]\par
\textbf{Action:} heat mug 1 with microwave 1\par
\smallskip\textcolor{CreditOrange}{\textbf{[Step 6]}}\par
\textbf{Observation:} You heat the mug 1 using the microwave 1.\par
\textbf{Reasoning:} [...] The microwave was then successfully used to heat mug 1. The next required action is to put this heated mug into coffeemachine 1. [...]\par
\textbf{Action:} go to coffeemachine 1\par
\smallskip\textcolor{CreditOrange}{\textbf{[Step 7]}}\par
\textbf{Observation:} You arrive at coffeemachine 1. On the coffeemachine 1, you see nothing.\par
\textbf{Reasoning:} [...] The coffeemachine was reached and confirmed empty, so it is a logical target for the heated mug.\par
\textbf{Action:} move mug 1 to coffeemachine 1\par
\smallskip\textbf{[Final environment feedback]}\par
You move the mug 1 to the coffeemachine 1.\par
\textbf{Recorded outcome:} SUCCEEDED; reward = 10.0; done = true.
\end{promptpanel}
\caption{\textbf{ALFWorld training trajectory.} The Actor locates a mug, heats it, and places it in the coffeemachine. The environment confirms completion after seven actions.}
\label{fig:trajectory-alfworld}
\end{figure}

\begin{figure}[H]
\begin{promptpanel}{WebShop trajectory (3B; training snapshot 120)}
\small
\textbf{Task:} Find me women's dresses with polyester spandex for daily wear with color: red, and size: large, and price lower than 40.00 dollars.\par
\smallskip\textcolor{CreditOrange}{\textbf{[Step 1]}}\par
\textbf{Observation:} 'Search'\par
\textbf{Reasoning:} [...] Let's open a search first and refine the query: `women's polyester/spandex dress daily red large under \$40'. This generates a focused search and is the appropriate next action.\par
\textbf{Action:} search[women's polyester/spandex dress daily red large under \$40]\par
\smallskip\textcolor{CreditOrange}{\textbf{[Step 2]}}\par
\textbf{Observation:} [...] 'Page 1 (Total results: 50)' [...] 'B08ZRR3DZT' [SEP] 'SheIn Women's Sexy One Shoulder Sleeveless Cutout Mini Club Pencil Bodycon Dress' [SEP] '\$19.99 to \$27.99' [...]\par
\textbf{Reasoning:} [...] This confirms a promising candidate, whose price range is below \$40. The click needs to open this product page before making further judgments.\par
\textbf{Action:} click[b08zrr3dzt]\par
\smallskip\textcolor{CreditOrange}{\textbf{[Step 3]}}\par
\textbf{Observation:} [...] 'size' [SEP] [...] 'large' [SEP] [...] 'color' [SEP] [...] 'red' [SEP] [...] 'Price: \$19.99 to \$27.99' [...] 'Buy Now'\par
\textbf{Reasoning:} [...] The admissible actions mention red and large, so one of those is still appropriate.\par
\textbf{Action:} click[large]\par
\smallskip\textcolor{CreditOrange}{\textbf{[Step 4]}}\par
\textbf{Observation:} [...] 'size' [SEP] [...] 'large' [SEP] [...] 'color' [SEP] [...] 'red' [SEP] [...] 'Price: \$19.99 to \$27.99' [...] 'Buy Now'\par
\textbf{Reasoning:} [...] It needs to select large and red as required variants before checkout. The only action taken was choosing large, so only action options are available.\par
\textbf{Action:} click[red]\par
\smallskip\textcolor{CreditOrange}{\textbf{[Step 5]}}\par
\textbf{Observation:} [...] 'size' [SEP] [...] 'large' [SEP] [...] 'color' [SEP] [...] 'red' [SEP] [...] 'Price: \$19.99 to \$27.99' [...] 'Buy Now'\par
\textbf{Reasoning:} You have already selected Large and Red on the SheIn dress, and the agent checks the displayed range of \$19.99--\$27.99, which is below \$40. [...] The observation contains Large and Red elements, so it supports the size and color selection, but the displayed price needs further confirmation.\par
\textbf{Action:} click[buy now]\par
\smallskip\textbf{[Final environment feedback]}\par
Thank you for shopping with us! [...] Purchased [SEP] asin [SEP] B08ZRR3DZT [SEP] options [SEP] \{"color": "red", "size": "large"\} [...] Your score (min 0.0, max 1.0) [SEP] 1.0 [...]\par
\textbf{Recorded outcome:} SUCCEEDED; reward = 10.0; done = true.
\end{promptpanel}
\caption{\textbf{WebShop training trajectory.} The Actor searches for a dress, opens a product, selects large and red, and purchases it. The terminal feedback records the selected options and a score of 1.0.}
\label{fig:trajectory-webshop}
\end{figure}

\subsection{Token masks and joint optimization}
Command-content masks select the parsed command span, excluding reasoning, delimiter tags, observations, and padding. The primary policy loss and regularization use all valid generated response tokens. Teacher scoring retains the Actor's sampled reasoning prefix. Only command tokens receive the direct OPSD term. The mean command gap $\bar d_{gjt}$ determines a shared modulation multiplier, which scales each command token's own gap $\delta_{gjt\ell}$, with the same token-mean loss reduction.

Analyzer credit uses raw summed command gaps $d_{gjt}$ before modulation. Equation~\ref{eq:rsloss} gives equal total weight to reflection and skill generation and averages each generated sequence over its own valid tokens. Zero-credit groups remain in the fixed group denominator. Actor and Analyzer gradients share the scheduled AdamW updates, with no additional R/S optimizer steps.

\Needspace{7\baselineskip}
\section{Baselines, benchmarks, and evaluation setting}
\label{app:evaluation}
\subsection{Baseline details}
\label{app:baselines}
Table~\ref{tab:maincomparison} compares Qwen2.5-3B-Instruct and Qwen2.5-7B-Instruct models. We summarize the baseline variants reported by AlignOPSD~\cite{alignopsd}. An asterisk denotes retrieved-skill access at inference; unmarked methods receive only the task prompt and interaction history.

\begin{itemize}[itemsep=3pt,topsep=4pt]
\item \textbf{Vanilla.} The instruction-tuned backbone acts directly in the environment without additional task-specific training.
\item \textbf{Skill-Prompt*.} A retrieved task-relevant skill is added to Vanilla's context at validation and test time.
\item \textbf{GRPO}~\cite{deepseekmath}. Terminal rewards are normalized within each task group to obtain trajectory advantages, which supervise the sampled response tokens without a learned critic.
\item \textbf{Skill-GRPO / Skill-GRPO*.} Both variants condition GRPO training on retrieved skills. Evaluation removes the skills for Skill-GRPO and retains them for Skill-GRPO*.
\item \textbf{OPSD}~\cite{opsd}. A privileged-context Teacher re-scores tokens from student-generated trajectories, providing detached token-level supervision. The student learns to reproduce these preferences using its ordinary interaction context.
\item \textbf{GRPO+OPSD.} A joint objective combines terminal-reward GRPO advantages with token-level supervision from the privileged-context Teacher.
\item \textbf{Skill-SD}~\cite{skillsd}. Skills condition the Teacher alone, and importance-weighted distillation transfers its guidance to the acting policy.
\item \textbf{RLSD.} Teacher--Student information produces a bounded scaling coefficient for GRPO updates. The outcome advantage determines the direction of each scaled update.
\item \textbf{SDAR}~\cite{sdar}. A gate regulates an auxiliary self-distillation loss alongside GRPO, while the original environmental advantage estimate is retained.
\item \textbf{AlignOPSD}~\cite{alignopsd}. Cross-rollout decision correspondence rectifies Teacher supervision, while adaptive decision spans and hierarchical credit assignment allocate the resulting guidance.
\item \textbf{Ours.} Task-specific R/S SFT initializes a shared Actor and Analyzer. Environmental command advantages calibrate signed Teacher preferences on executable command tokens. State-centered credit trains reflections and skills, linking Teacher construction to subsequent policy updates. Inference uses the trained Actor with the ordinary task prompt and interaction history.
\end{itemize}

The published baseline values are taken from AlignOPSD Table~1~\cite{alignopsd}. Appendix~\ref{app:implementation} specifies our initialization and training settings; the following subsections describe our task panels, inference settings, and reporting conventions.

\subsection{Task panels and metrics}
ALFWorld evaluation covers all 140 seen and 134 unseen tasks. Table~\ref{tab:categorycounts} gives the six category sizes. If $k_c$ of $n_c$ tasks succeed in category $c$, macro success is $\frac{100}{6}\sum_{c=1}^{6}k_c/n_c$. Each category therefore contributes equally. Success is determined by the first terminal environment outcome.

\begin{table}[H]
\centering\small
\caption{ALFWorld evaluation tasks by category.}
\label{tab:categorycounts}
\begin{tabularx}{\linewidth}{@{}Xrrrrrrr@{}}
\toprule
Split & Pick & Look & Clean & Heat & Cool & Pick2 & Total \\
\midrule
Seen & 35 & 13 & 27 & 16 & 25 & 24 & 140 \\
Unseen & 24 & 18 & 31 & 23 & 21 & 17 & 134 \\
\bottomrule
\end{tabularx}
\end{table}

WebShop uses 128 goals sampled without replacement. Our repeated evaluations draw a 128-goal panel for each evaluation seed; methods using the same seed share the panel. The catalogue seed remains 0, with the training-compatible search index of 1,000 products. Success is the percentage of tasks completed fully. Score is the mean normalized terminal environment reward multiplied by 100, capturing partial satisfaction of the requested product attributes.

\subsection{Inference settings}
Evaluation uses temperature $0.4$, top-$p=1$, top-$k=-1$, and one trajectory per task. The Actor receives the ordinary task prompt and interaction history, without appended skills or online R/S generation. ALFWorld allows at most 30 actions with history length five; WebShop allows at most 15 actions with history length two. Prompt and response limits follow Appendix~\ref{app:implementation}. Infrastructure errors are recorded separately from completed task failures.

\subsection{Reporting conventions}
All task-performance results for our method and its ablations use the final checkpoint after 150 online updates. The 134-task unseen split is used only for final evaluation.

All training runs use a fixed random seed.
After training, each checkpoint is evaluated under three evaluation seeds, and we report the mean and standard deviation in Tables~\ref{tab:maincomparison} and~\ref{tab:bestunseen}.

The local GRPO baselines in Table~\ref{tab:bestunseen} use the same initialization, task sampling, rollout budget, Actor optimization settings, and evaluation protocol as the full method in Table~\ref{tab:online-settings}. Their online updates optimize $\mathcal L_{\mathrm{GRPO}}$ alone, including the same reference-policy KL penalty and entropy bonus, for 150 updates. They are evaluated on the same unseen task panel with the same inference settings as our models.

The GRPO row in Table~\ref{tab:webshopablation} repeats the published baseline values quoted in Table~\ref{tab:maincomparison}. Published baseline numbers come from Table~1 of AlignOPSD~\cite{alignopsd} and retain their source aggregation.

\subsection{Ablation configurations and evaluation}
\label{app:ablationbranches}
Table~\ref{tab:ablation-configurations} specifies the full method and the three 3B ablation variants. All variants generate eight reflections and one skill per task group during online training. Variant (a) omits R/S SFT and starts from Qwen2.5-3B-Instruct. Variant (b) extends direct Teacher supervision to the complete response, including reasoning. Variant (c) retains command-only supervision and online R/S generation in both environments; it sets the modulation multiplier to one and removes the direct R/S objective. The remaining settings follow Table~\ref{tab:online-settings}.

\paragraph{Full-response supervision in variant (b).}
At each interaction turn, we compute the modulation multiplier $m_{gjt}$ from the executable command's mean gap $\bar d_{gjt}$ and environmental advantage $\widehat A_{gjt}$ using Equation~\ref{eq:modulation}. This multiplier is broadcast to all valid generated tokens in the same Actor response, including reasoning and command content. Each token retains its own Teacher--Actor gap, giving
\begin{equation}
    w^{(b)}_{gjt\ell}
    =q_g\,m_{gjt}\,\delta_{gjt\ell},
    \qquad (g,j,t,\ell)\in\mathcal B_A.
    \label{eq:fullresponse-ablation-weight}
\end{equation}
The loss uses the clipped token objective in Equation~\ref{eq:teacherloss}. Both supervision scopes use the Actor token-mean denominator $|\mathcal B_A|$ in Equation~\ref{eq:opsd-objective}, Teacher coefficient $\beta_T=0.01$, and PPO clipping ratio $0.2$. The full-response branch applies these token-level Teacher terms throughout the sampled response under the command-derived multiplier.

\begin{table}[htbp]
\centering\small
\setlength{\tabcolsep}{3pt}
\caption{Training components used in the ablation study. Scope denotes the tokens covered by direct Teacher supervision. Online R/S denotes reflection and skill generation during training.}
\label{tab:ablation-configurations}
\begin{tabular*}{\linewidth}{@{\extracolsep{\fill}}lccccc@{}}
\toprule
Variant & SFT & Scope & Mod. & \shortstack{R/S\\loss} & \shortstack{Online\\R/S} \\
\midrule
Full & Yes & Command & Yes & Yes & Yes \\
(a) & No & Command & Yes & Yes & Yes \\
(b) & Yes & Response & Yes & Yes & Yes \\
(c) & Yes & Command & No & No & Yes \\
\bottomrule
\end{tabular*}
\end{table}

Each branch trains for 150 online updates and is evaluated at its final checkpoint. ALFWorld evaluation uses all 140 seen tasks, at most 30 actions, and history length five. WebShop evaluation uses the fixed 128-task panel, at most 15 actions, and history length two. Both environments use temperature $0.4$ and one trajectory per task. We report six-category macro success for ALFWorld and Score and Success for WebShop. The GRPO row contains the published baseline values described above.

\section{Mechanistic analysis and training dynamics}
\label{app:extended}
\label{app:mechanismdiagnostics}
\subsection{Paired measurement of Teacher weight}
The analysis in Figure~\ref{fig:mechanism}(a,b) uses recorded command spans, likelihood gaps, modulation weights, and primary policy advantages from the two 3B runs specified in Appendix~\ref{app:online-settings}. We recorded nine snapshots per environment. Duplicate padding rows are removed, and each executed command contributes one observation. We retain commands with an active Teacher span and a primary advantage satisfying $|\widehat A_i^{\mathrm{RL}}|>10^{-8}$. The same commands are used before and after modulation.

The raw command coefficient is the summed Teacher--Actor gap $d_i$. The final command coefficient is the mean saved coefficient over the command's content tokens. These command-level values enter the reward-alignment statistic in Section~\ref{sec:mechanism}. Each condition is normalized by its own total absolute coefficient mass. This comparison includes both command-length normalization and environmental modulation. The shaded region in Figure~\ref{fig:mechanism}(a,b) shows the paired percentage-point difference.

The reference advantage $\widehat A_i^{\mathrm{RL}}$ comes from the primary policy objective: the recorded outcome includes the configured invalid-command penalty and is standardized within the task group. The auxiliary graph advantage $\widehat A_i$ determines modulation. The primary advantage provides the outcome-derived reference for the measurement, and the auxiliary advantage controls the Teacher multiplier.

\subsection{Return concordance of skill-induced preferences}
For each skill, we pair distinct commands from different trajectories when their stored Actor prompts and admissible-action sets match exactly. Teacher scoring retains the reasoning prefix sampled in each response. Let $\Delta d_{ij}=d_i-d_j$ be the difference in raw command gaps and $\Delta R_{ij}=R_i-R_j$ the difference in recorded returns-to-go. Pairs with $|\Delta R_{ij}|\leq10^{-8}$ are omitted. With $\varepsilon=10^{-8}$, the pair score is
\begin{equation}
 h_{ij}=\begin{cases}
 \frac{1}{2}, & |\Delta d_{ij}|\leq\varepsilon,\\
 1, & |\Delta d_{ij}|>\varepsilon\ \text{and}\ \Delta d_{ij}\Delta R_{ij}>0,\\
 0, & \text{otherwise}.
 \end{cases}
\end{equation}
Concordance is $100$ times the mean pair score within a skill. Skills with at least one eligible pair enter the analysis. We then give each skill equal weight, so skills with more observed command pairs do not dominate the comparison.

The pooled analysis uses the saved snapshots after the initial update. Within each task, the median bounded R/S credit $c_g$ separates the higher- and lower-credit groups; median ties belong to the lower-credit group. Figure~\ref{fig:mechanism}(c,d) displays individual skills, group means, and interquartile ranges. The complete measurements underlying the figure accompany the supplementary data.

\Needspace{9\baselineskip}
\subsection{Training dynamics}
\label{app:training-dynamics}
\begin{figure}[htbp]
\centering
\includegraphics[width=\linewidth]{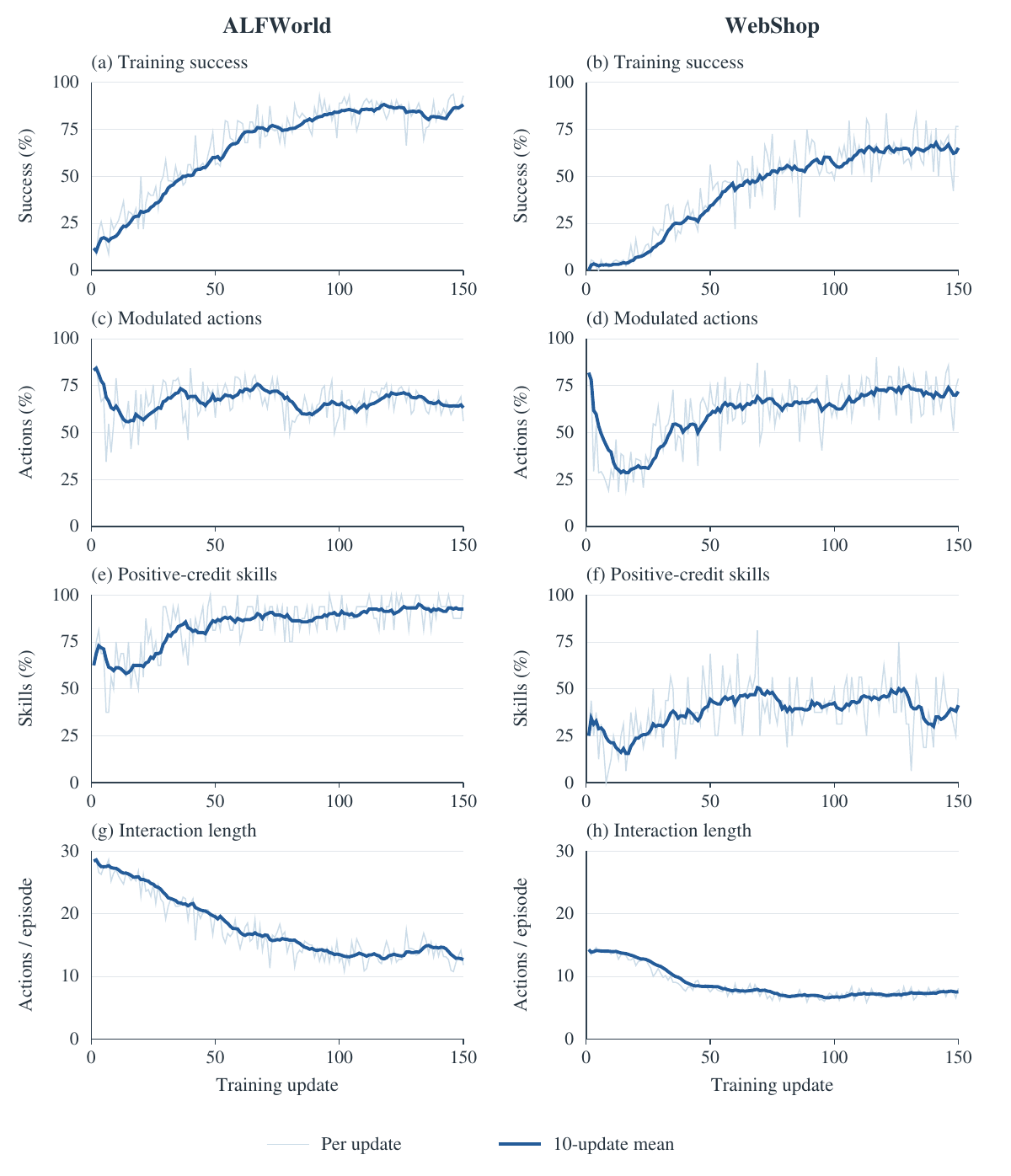}
\caption{\textbf{Training dynamics of the 3B agents.} Columns show ALFWorld and WebShop; rows show training success, modulated-action coverage, the fraction of skills receiving positive R/S credit, and mean interaction length. Light curves show individual updates; dark curves show trailing ten-update means, using available updates at the start. Success rises and episodes shorten while environmental feedback continues to select and modulate supervision.}
\label{fig:training3b}
\end{figure}

\begin{figure}[htbp]
\centering
\includegraphics[width=\linewidth]{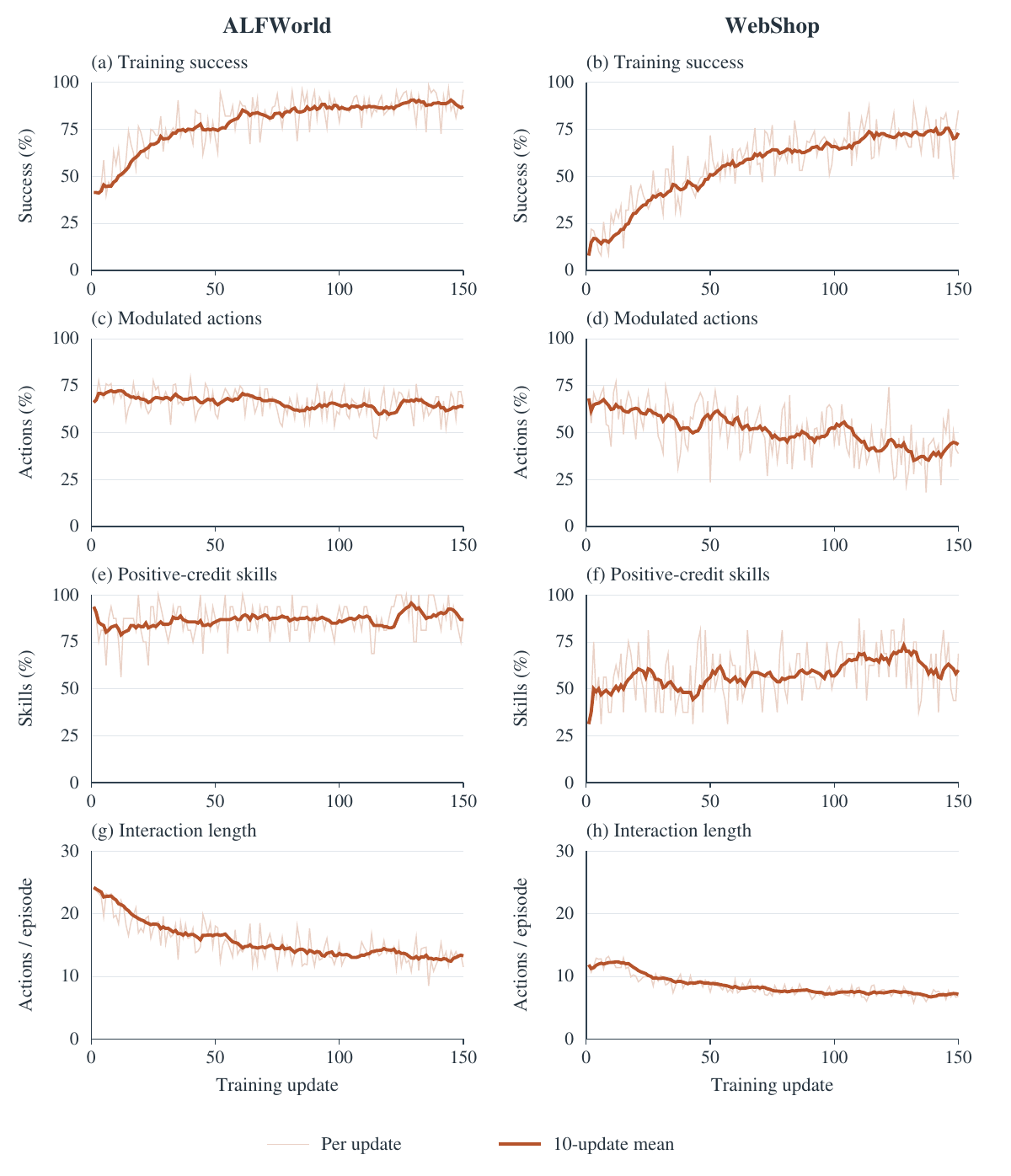}
\caption{\textbf{Training dynamics of the 7B agents.} Columns and rows follow Figure~\ref{fig:training3b}, with the same axis ranges and moving-average convention. Both tasks show increasing success and shorter interactions. Modulation coverage remains active on ALFWorld and declines on WebShop.}
\label{fig:training7b}
\end{figure}

Figures~\ref{fig:training3b} and~\ref{fig:training7b} show the 3B and 7B command-only runs over 150 updates. Success and interaction length summarize 128 trajectories per update; positive-credit fractions count all sixteen skills, including zero-credit groups. Modulation coverage retains the logged training-batch rows and active-Teacher mask.

\end{document}